\documentclass[10pt]{article}
\usepackage[preprint]{tmlr}

\usepackage{amsmath,amsfonts,bm}

\def\eqref#1{equation~\ref{#1}}

\def\1{\bm{1}}

\DeclareMathAlphabet{\mathsfit}{\encodingdefault}{\sfdefault}{m}{sl}
\SetMathAlphabet{\mathsfit}{bold}{\encodingdefault}{\sfdefault}{bx}{n}

\usepackage{hyperref}
\usepackage{url}
\usepackage{booktabs}
\usepackage{graphicx}
\usepackage{multirow}
\usepackage{placeins}
\usepackage{amsmath,amssymb}
\usepackage{xcolor}
\usepackage[capitalize]{cleveref}

\title{Scout Before You Route: Predicting Task Success and \\ Token Count from a Small LLM Prefill}
\author{\name Aristides Milios \email aristides.milios@mila.quebec \\
\addr Mila -- Quebec AI Institute, Universit\'e de Montr\'eal \\
\addr ServiceNow AI Research
\AND
\name Christopher Pal \email christopher.pal@servicenow.com \\
\addr Mila -- Quebec AI Institute, Polytechnique Montr\'eal \\
\addr ServiceNow AI Research \\
\addr Canada CIFAR AI Chair}

\begin{document}
\maketitle
\lhead{}
\begin{abstract}
    Routers typically choose a model per query by trading predicted correctness against cost. Most price each model at a fixed cost, as if every query used the same number of output tokens, but the output length of reasoning models varies by an order of magnitude across queries. When is per-query cost worth predicting? What factors drive cost? How cheaply can predictions be made? We predict \textbf{both} the success probability \textbf{and} the output length of candidate models from the prefill activations of a single small frozen LLM, the \emph{scout}. With reasoning candidates, perfect per-query cost knowledge would save 29--61\% of spending on eight of nine datasets, and much less with non-reasoning candidates on the same problems. On those eight datasets, our predicted costs save 9--33\% at matched accuracy relative to median-length pricing, up to 69\% of what perfect knowledge would save. They are significantly cheaper than each estimator we reimplemented or adapted from published work, including ZeroRouter and a MixLLM-style predictor, on a majority of datasets. Output length depends mostly on the problem. On most math and code datasets, harder problems are longer, so predicted success alone prices cost nearly as well. On knowledge and mixed-format datasets they are not, and predicted success falls well behind a probe trained on length. The probe keeps most of its saving with 20 labelled problems and transfers without labels to four of seven new benchmarks. On a mixture of all nine benchmarks, it saves 22\% beyond pricing each benchmark at its own median length. Savings stop growing at about a 2B scout, so above 4B a larger scout costs more than it adds.
\end{abstract}

\section{Introduction}
Routing reasoning models requires predicting both correctness and cost before generation. A model's price per token is known in advance, but the number of tokens it will write for a given problem is not. A model that is cheap on average can be expensive on a particular problem: on LiveCodeBench, a single model's output length differs 11--121$\times$ between its 10th- and 90th-percentile problems (\cref{tab:poolstats}).

\begin{figure}[t]
\centering
\includegraphics[width=\linewidth]{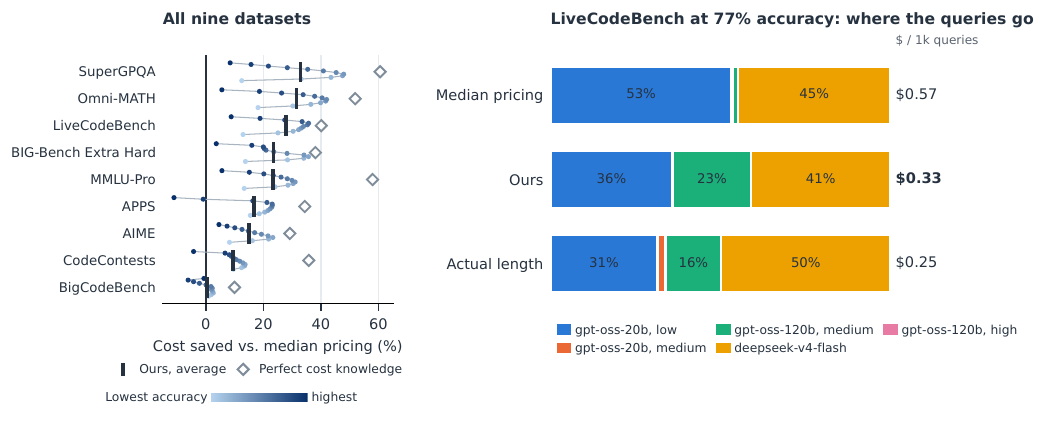}
\caption{\textbf{What predicting cost saves, and where the queries go.} Left: cost saved relative to median pricing on all nine datasets at each of 12 matched accuracies, from the lowest (light, low in the row) to the highest (dark, high in the row) in the range both routers reach, averaged over ten training sets. Ticks mark our average over the range and diamonds the same average for perfect per-query cost knowledge (\cref{tab:pools} gives 95\% intervals). Right: share of LiveCodeBench test queries sent to each candidate at 77\% accuracy by three routers that price a query by median length, by our cost probes, or by its actual length (perfect knowledge), with the spend per 1,000 queries (one training set). Ours moves queries from gpt-oss-20b at low effort to gpt-oss-120b at medium effort.}
\label{fig:hero}
\end{figure}

Prefill routers predict correctness from a small encoder's prompt activations; the prefill router of Varshney et al.~\citep{prefill} explicitly prices outputs at median training length. We predict output length from the same activations of a small frozen \emph{scout} model, which only processes the prompt and generates nothing. The scout is in effect an \emph{actuarial scout}: as an actuary prices each policy from the applicant rather than charging everyone the average premium, it prices each query before any model is called. Cost prediction adds one linear probe per candidate to an existing prefill router, without another encoder pass. Most routers that do predict per-query cost evaluate it only inside their full system; the one that isolates it finds that it adds little~\citep{carrot}, on pools with almost no reasoning models. We isolate it on reasoning candidates, against both constant pricing and perfect per-query cost knowledge.

This raises three questions:
\begin{enumerate}
\item When is per-query cost worth predicting?
\item What factors drive cost for a given input?
\item How cheaply can a router obtain a cost prediction: with how many labels, and with a prefill pass from how small an LLM?
\end{enumerate}

We contribute:
\begin{enumerate}
\item a measurement of what perfect knowledge of per-query cost saves: 29--61\% of spending at matched accuracy on eight of nine datasets with reasoning candidates, and much less with non-reasoning candidates on the same problems (\cref{sec:headroom});
\item a prefill cost probe that saves 9.4--33.0\% relative to median pricing on eight of nine datasets, up to 69\% of what perfect knowledge would save. Against external estimators it is ahead in 69 of 72 comparisons and significantly ahead in 50. It is significantly ahead of all of them on LiveCodeBench, MMLU-Pro, AIME and SuperGPQA, and of all but one on Omni-MATH (\cref{fig:hero,tab:allest});
\item a mechanism: length depends mostly on the problem (model lengths co-vary, and models output relatively consistent lengths across samples on the same query). Where length correlates with difficulty, length from predicted success suffices. Where it does not, a predictor trained on length is needed, and on MMLU-Pro only the cost probe captures it. Our two failures come from too little headroom, or headroom no estimator we tried can capture (\cref{sec:mechanism,sec:when});
\item generalisation: the cost probe keeps most of its saving with 20 labelled problems and for a new model onboarded from 10 examples; it transfers with no target labels to four of seven benchmarks; and in mixed traffic it saves 22.4\% beyond pricing by benchmark identity (\cref{sec:general});
\item scout size and controls: savings do not grow with the scout from 1.7B to 14B, so above 4B a larger scout costs more than it adds, while a text-embedding model of the same size and fully fine-tuned small models predict cost worse (\cref{sec:general,sec:ablations}).
\end{enumerate}

\paragraph{Related work.}
Learned routers pick one model per query from the prompt, trading predicted quality against price: RouteLLM learns a strong-versus-weak decision from preference data~\citep{routellm}, RouterBench standardizes the evaluation~\citep{routerbench}, and IRT-Router models query difficulty and model ability with item response theory~\citep{irtrouter}. Closest to us, the prefill router of Varshney et al.\ predicts each model's correctness from a small encoder's prompt activations~\citep{prefill}, and pre-generation activations are known to encode whether a model will fail~\citep{encodefailures}. These routers, like SWE-Router for agentic software engineering~\citep{swerouter}, price each model at a per-model constant; the prefill router uses the median training output length, which is our main reference.

Fewer routers predict per-query cost: MixLLM regresses output length on query embeddings~\citep{mixllm}, CARROT uses embedding nearest neighbours or a fine-tuned RoBERTa~\citep{carrot}, GraphRouter predicts cost on a query--model graph~\citep{graphrouter}, Route-To-Reason predicts reasoning-model length from text embeddings~\citep{routetoreason}, ZeroRouter prices queries by item-response difficulty bins~\citep{zerorouter}, and SCOPE estimates a query's cost from retrieved past behaviour~\citep{scope}. R2-Router routes over pairs of a model and an output-length budget, so length is controlled rather than predicted~\citep{r2router}, and SCX Router predicts task type, difficulty and an expected output length from a small decoder's KV cache as properties of the task, not of each candidate model~\citep{scx}. Activation-based length prediction also supports serving and scheduling of a single model~\citep{alps,egtp,outlets}; we predict the lengths of other models from one shared scout. Of MixLLM, CARROT and Route-To-Reason, only CARROT isolates its cost predictor, against the same router pricing each model at its training-average cost, and it finds that per-query cost adds little on RouterBench and SPROUT~\citep{carrot}; none of the three compares with perfect per-query cost knowledge. We compare cost estimators directly, including reimplementations of these and ZeroRouter, with every estimator given the same success predictions and measured against both constant pricing and perfect cost knowledge.

Recent large-scale evaluations find that routers land close to an oracle that assigns each dataset its best model, and attribute much of their gain to coarse domain structure~\citep{llmrouterbench}; a static assignment by task type closes most of the gap to the per-query oracle on a small benchmark~\citep{tasktype}. Both concern which model answers correctly; we measure what per-query cost is worth beyond the benchmark a query comes from.

\begin{figure}[t]
\centering
\includegraphics[width=\linewidth]{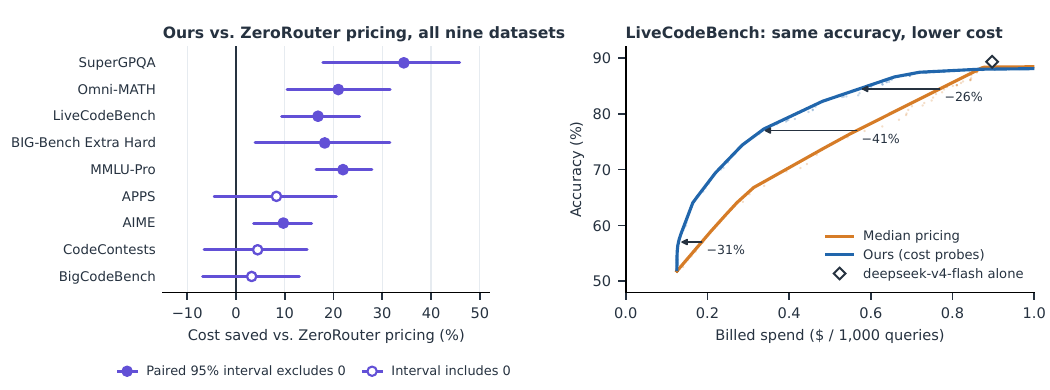}
\caption{\textbf{Cost saved at matched accuracy, billed prices.} Left: cost saved by ours relative to ZeroRouter's pricing, both with our success predictions and our 4B scout, with 95\% intervals; hollow where the interval includes zero (\cref{tab:allest}). Right: LiveCodeBench cost--accuracy frontiers of our cost probe and median pricing (one training set); arrows mark the saving at matched accuracy, and the diamond is deepseek-v4-flash on every query.}
\label{fig:overview}
\end{figure}

\section{Cost Prediction and Routing}
For query $x$ and candidate $m$ (a model at a fixed reasoning effort, served by an endpoint), let $p_m(x)$ be the probability of a correct answer and $c_m(x)$ the expected generation cost. One candidate is chosen before generation,
\begin{equation}
m^*(x;V)=\arg\max_m\{V\hat p_m(x)-\hat c_m(x)\},
\label{eq:route}
\end{equation}
where $V$ is the value of a correct answer; sweeping $V$ trades accuracy for cost. Every query is answered once, with no verifier or fallback.

\paragraph{Probes.}
We concatenate mean- and last-token activations from eight layers of a frozen Qwen3-4B~\citep{qwen3} and fit one linear probe~\citep{alain2016probes} per candidate and target: for success an L2-regularized logistic regression, calibrated by Platt scaling~\citep{platt1999}; for cost a ridge regression of log mean output length, corrected for the bias of exponentiating by Duan's smearing~\citep{duan1983}. The predicted length is priced at the candidate's token rates, $\hat c_m(x)=r_m^{\rm in}n^{\rm in}_m(x)+r_m^{\rm out}\hat\ell_m(x)$. \Cref{app:protocol} gives the features, regularization and calibration.

\section{Experimental Protocol}\label{sec:protocol}
The main pool has five candidates, called through OpenRouter: gpt-oss-20b at low and medium effort~\citep{gptoss}, deepseek-v4-flash~\citep{deepseekv4}, and gpt-oss-120b at medium and high effort. Endpoints of the same open-weight model differ across providers in accuracy, length and price~\citep{samemodel,gateways}. We pin deepseek-v4-flash to one provider because the length it produced varied significantly between providers, while that of gpt-oss did not; the gpt-oss candidates are served by whichever provider OpenRouter selects. We use nine datasets: LiveCodeBench~\citep{lcb}, Omni-MATH~\citep{omni}, MMLU-Pro~\citep{mmlupro}, AIME~\citep{aime}, APPS~\citep{apps}, BigCodeBench~\citep{bigcodebench}, CodeContests~\citep{codecontests}, SuperGPQA~\citep{supergpqa} and BIG-Bench Extra Hard~\citep{bbeh}. Each is split into training, calibration and test problems (\cref{app:pools,app:protocol}). On three of the datasets we also extend the pool with three reasoning models from other families (\cref{app:families}). \Cref{tab:routes} gives each candidate's accuracy and cost on every test set.

\paragraph{Prices.}
Each model has one price per input and per output token: its average billed price, a least-squares fit of the billed cost of our calls on their tokens (USD per million output tokens: 0.13 for gpt-oss-20b, 0.08 for deepseek-v4-flash, 0.26 for gpt-oss-120b; for the gpt-oss models, an average over the providers OpenRouter chose). A call's cost is its tokens at these prices; on the Omni-MATH and MMLU-Pro test sets, where we recorded each call's billed cost, we use that instead. Costing those two test sets as tokens, or fitting the prices on training calls only, moves our saving relative to median pricing by at most 1.0 point and changes the sign or significance of no comparison in \cref{tab:main} (\cref{app:protocol}).

\paragraph{Evaluation.}
Every estimator is fitted on training problems only and evaluated on the test set, with the same success predictions, apart from one ZeroRouter row that uses its own. Sweeping $V$ gives one policy per value and a (cost, accuracy) point on the test set; an estimator's frontier is the upper concave hull of its points, each of which is achievable by sending each query to one of two policies at random with fixed probabilities~\citep{routerbench}. We compare two cost estimators $a,b$ at 12 accuracies that both frontiers reach and report one minus the geometric mean of $C_a/C_b$, the cost saved by $a$ at matched accuracy. Pricing each problem at its actual output length (perfect cost knowledge) is the practical upper bound for every estimator; its saving over median pricing is the \emph{headroom}, and our share of it the \emph{capture}. Every result is a mean over ten \emph{training sets}, each of which reshuffles the training and calibration problems, keeps the test set and refits everything; intervals are paired bootstrap intervals over test problems, pooled over the ten. Analyses marked ``one training set'' use one split (\cref{app:protocol}). \Cref{sec:deploy} checks the results without interpolation, with $V$ fixed in advance on calibration.

\begin{table}[t]
\centering
\footnotesize
\setlength{\tabcolsep}{3.5pt}
\begin{tabular}{lrrrrrrrrr}
\toprule
Cost saved by ours vs. & LCB & Omni & MMLU-Pro & AIME & APPS & BCB & CC & SuperGPQA & BBEH \\
\midrule
\emph{External estimators} \\
Median pricing & \textbf{27.8} & \textbf{31.5} & \textbf{23.4} & \textbf{14.9} & \textbf{16.8} & 0.4 & \textbf{9.4} & \textbf{33.0} & \textbf{23.5} \\
Mean pricing & \textbf{27.8} & \textbf{34.3} & \textbf{25.8} & \textbf{15.0} & \textbf{13.3} & 3.8 & \textbf{9.4} & \textbf{35.8} & \textbf{22.7} \\
Prompt-feature GBM & \textbf{20.7} & \textbf{25.8} & \textbf{20.0} & \textbf{14.2} & 9.7 & 1.4 & 8.0 & \textbf{19.1} & 2.1 \\
MixLLM-style embeddings & \textbf{16.2} & \textbf{24.7} & \textbf{15.5} & \textbf{13.5} & \textbf{16.4} & 1.7 & 4.3 & \textbf{15.7} & $-$1.3 \\
CARROT-style $k$NN & \textbf{17.5} & \textbf{23.2} & \textbf{14.3} & \textbf{15.0} & \textbf{15.3} & 0.4 & 4.0 & \textbf{20.9} & 0.3 \\
Route-To-Reason & \textbf{17.7} & \textbf{23.2} & \textbf{14.8} & \textbf{13.9} & \textbf{13.4} & 2.9 & 5.0 & \textbf{15.1} & 4.5 \\
ZeroRouter pricing, our success & \textbf{16.8} & \textbf{21.0} & \textbf{22.0} & \textbf{9.7} & 8.3 & 3.2 & 4.5 & \textbf{34.4} & \textbf{18.2} \\
ZeroRouter pricing, its success & \textbf{19.3} & 14.2 & \textbf{18.4} & \textbf{8.3} & 2.3 & $-$10.7 & $-$1.0 & \textbf{35.2} & \textbf{24.5} \\
\midrule
\emph{Our ablations} \\
Length from predicted success & 3.2 & $-$3.0 & \textbf{18.4} & 4.3 & $-$2.2 & 2.4 & $-$0.8 & 15.3 & \textbf{22.6} \\
One length per query (SCX-style) & \textbf{10.7} & \textbf{21.1} & \textbf{10.4} & \textbf{8.8} & 5.7 & 3.0 & 0.6 & 5.7 & 5.5 \\
\bottomrule
\end{tabular}
\caption{\textbf{Cost saved (\%) by our cost probe relative to each estimator} at matched accuracy. Every row uses our success predictions except the second ZeroRouter row, which uses its own. Both ZeroRouter rows predict query difficulty from our 4B scout's activations (\cref{app:baselines} gives the version as published). \textbf{Bold}: statistically significant (95\% paired bootstrap interval excludes zero). Ours is ahead in 69 of the 72 comparisons with external estimators, significantly in 50. Estimators are described in \cref{app:baselines}.}
\label{tab:allest}
\end{table}

\begin{table}[t]
\centering
\footnotesize
\setlength{\tabcolsep}{2.6pt}
\begin{tabular}{lrrrrrrrrrrrrrrrrrr}
\toprule
 & \multicolumn{2}{c}{LCB} & \multicolumn{2}{c}{Omni} & \multicolumn{2}{c}{MMLU-Pro} & \multicolumn{2}{c}{AIME} & \multicolumn{2}{c}{APPS} & \multicolumn{2}{c}{BCB} & \multicolumn{2}{c}{CC} & \multicolumn{2}{c}{SGPQA} & \multicolumn{2}{c}{BBEH} \\
\cmidrule(lr){2-3}\cmidrule(lr){4-5}\cmidrule(lr){6-7}\cmidrule(lr){8-9}\cmidrule(lr){10-11}\cmidrule(lr){12-13}\cmidrule(lr){14-15}\cmidrule(lr){16-17}\cmidrule(lr){18-19}
Candidate & Acc. & \$ & Acc. & \$ & Acc. & \$ & Acc. & \$ & Acc. & \$ & Acc. & \$ & Acc. & \$ & Acc. & \$ & Acc. & \$ \\
\midrule
20b, low & 51.8 & 0.13 & 41.9 & 0.21 & 62.1 & 0.05 & 57.2 & 0.26 & 58.4 & 0.12 & 45.3 & 0.06 & 49.1 & 0.14 & 31.0 & 0.05 & 23.3 & 0.11 \\
20b, medium & 68.2 & 0.65 & 57.5 & 1.20 & 69.4 & 0.18 & 88.3 & 1.18 & 73.5 & 0.74 & 48.3 & 0.12 & 70.6 & 0.85 & 36.0 & 0.24 & 34.8 & 1.21 \\
deepseek-v4-flash & 89.3 & 0.90 & \textbf{71.3} & 1.66 & \textbf{86.2} & 0.33 & \textbf{98.3} & 0.86 & 84.1 & 1.05 & 52.1 & 0.11 & 84.6 & 1.15 & \textbf{67.9} & 0.47 & \textbf{64.7} & 1.54 \\
120b, medium & 82.1 & 0.64 & 60.3 & 1.12 & 77.2 & 0.23 & 89.5 & 1.23 & 79.3 & 0.72 & \textbf{53.4} & 0.18 & 78.4 & 0.84 & 46.0 & 0.27 & 45.3 & 1.45 \\
120b, high & \textbf{89.5} & 3.73 & 66.6 & 4.57 & 79.3 & 0.90 & 97.0 & 3.14 & \textbf{85.7} & 4.48 & 53.0 & 0.64 & \textbf{85.9} & 4.87 & 47.8 & 1.48 & 54.7 & 5.26 \\
\bottomrule
\end{tabular}
\caption{\textbf{Each candidate alone, without routing.} Accuracy (Acc., \%) and mean cost per call (\$, USD per 1,000 calls) when every test problem goes to one candidate (billed costs on Omni-MATH and MMLU-Pro, tokens at billed prices elsewhere). SGPQA: SuperGPQA. Bold: most accurate. Costs differ 11--49$\times$, and accuracy is not monotone in price: deepseek-v4-flash is the most accurate on five datasets and within 1.6 points of gpt-oss-120b at high effort on the rest, at 17--37\% of its cost.}
\label{tab:routes}
\end{table}

\section{How much is per-query cost worth?}\label{sec:headroom}
Knowing each query's actual cost is the practical upper bound for a cost estimator. \Cref{tab:pools} reports this headroom on all nine datasets, with our saving and the share of it we capture (\cref{sec:capture}).

\begin{table}[t]
\centering
\footnotesize
\setlength{\tabcolsep}{3.3pt}
\begin{tabular}{llrrrrr}
\toprule
Dataset & Domain & Test $n$ & Headroom & Ours vs.\ median & Capture & Ours vs.\ length \\
 & & & & & & from pred.\ success \\
\midrule
LiveCodeBench & code & 341 & 40.1 & 27.8 [13.9, 35.6] & .69 & 3.2 \\
Omni-MATH & math & 1,000 & 51.9 & 31.5 [23.9, 37.7] & .61 & $-3.0$ \\
MMLU-Pro & knowledge & 6,500 & 58.0 & 23.4 [17.1, 29.0] & .40 & \textbf{18.4} \\
AIME 1983--2024 & math & 281 & 29.2 & 14.9 [10.8, 20.1] & .51 & 4.3 \\
APPS & code & 377 & 34.4 & 16.8 [4.2, 30.4] & .49 & $-2.2$ \\
BigCodeBench & code & 421 & 10.0 & 0.4 [$-$9.5, 9.1] & .04 & 2.4 \\
CodeContests & code & 266 & 35.8 & 9.4 [0.4, 19.8] & .26 & $-0.8$ \\
SuperGPQA & knowledge & 300 & 60.6 & 33.0 [14.6, 45.7] & .54 & 15.3 \\
BIG-Bench Extra Hard & reasoning & 300 & 38.1 & 23.5 [9.4, 35.2] & .62 & \textbf{22.6} \\
\bottomrule
\end{tabular}
\caption{\textbf{Headroom and capture on every dataset.} Cost saved (\%) relative to median pricing by perfect per-query cost knowledge (headroom) and by our cost probe, with 95\% intervals; capture is their ratio. Last column: cost saved by the cost probe relative to our ablation that takes length from predicted success; bold: interval excludes zero.}
\label{tab:pools}
\end{table}

\paragraph{Headroom is large on eight of nine datasets.}
On eight of nine datasets perfect per-query cost knowledge would save 29--61\%; BigCodeBench has little to capture (10.0\%; \cref{sec:when}).

\paragraph{Per-query cost is worth less with non-reasoning candidates.}
We also collected five non-reasoning candidates on the same problems (deepseek-v4-flash with thinking disabled, Llama-3.1-8B~\citep{llama3}, Qwen3-30B-A3B-Instruct, Llama-3.3-70B and Qwen3-235B-A22B-Instruct) and fitted the probes as before (\cref{app:nonreason}). Headroom is 1.3--2.3$\times$ smaller than with reasoning candidates, our saving 2.3$\times$ (LCB) to about 30$\times$ (MMLU-Pro) smaller (\cref{tab:nonreason}), and no other estimator does better on Omni-MATH or MMLU-Pro. Much of the non-reasoning variation is degenerate output that the prompt predicts poorly, such as Llama-3.1-8B looping to its 16k cap on 9--12\% of math problems. With thinking on, deepseek-v4-flash gains 2.8--22.1 points of accuracy for 11--33$\times$ the output, and the problems that are long with thinking are long without it (log-length correlation .66--.76).

\section{How much of it our method captures}\label{sec:capture}
\paragraph{Against median pricing.}
At matched accuracy, our router spends significantly less than median pricing on eight of the nine datasets, from 9.4\% on CodeContests to 33.0\% on SuperGPQA (\cref{fig:hero,tab:pools}), and on seven of them the cost probe captures two fifths to over two thirds of the headroom (.40--.69). In absolute terms, on LiveCodeBench our router reaches 90\% of deepseek-v4-flash's accuracy at 49\% of what calling it on every query costs, against 69\% for median pricing (geometric mean over ten training sets). Perfect cost knowledge assumes that the random variation in each scored sample's length is known in advance, which no predictor can foresee, so these capture figures are conservative (\cref{app:headroom}). BigCodeBench, where the probe saves nothing, and CodeContests are covered in \cref{sec:when}.

\paragraph{We are ahead in 69 of 72 comparisons with external estimators.}
Ours is significantly ahead in 50 and significantly behind in none (\cref{fig:overview,tab:allest}), also on random re-splits that reshuffle the test problems (\cref{tab:resplit}). Of the other 22, one is ZeroRouter with its own success model on Omni-MATH (behind ours in 7 of 10 training sets), and the rest fall on four datasets. On BigCodeBench no estimator, not even perfect cost knowledge, is distinguishable from median pricing. On APPS and CodeContests ours is significantly cheaper than median pricing and no external estimator is, and three to six of them lie between the two, distinguishable from neither. On BIG-Bench Extra Hard four text-based estimators save as much as ours, all significantly more than median pricing (\cref{sec:mechanism}). On LCB, Omni-MATH and MMLU-Pro, every external estimator loses by 14--34 points, with intervals excluding zero except ZeroRouter with its own success model on Omni-MATH (\cref{tab:main}). Ours is significantly ahead of each in every training set there, except in that comparison.

\paragraph{Where the saving comes from.}
The saving is not only a price gap: with every candidate charged the same price per token and actual lengths kept, the cost probe still saves 13.3\% (LCB), 14.3\% (Omni-MATH) and 36.3\% (MMLU-Pro) relative to median pricing (one training set), and price gaps larger than the billed ones shrink both the headroom and the saving (\cref{app:subsets}). It needs a cheap candidate that solves a good share of problems: without gpt-oss-20b at low effort most of it disappears (LCB 27.8\% to 6.5\%, Omni-MATH 31.5\% to 16.2\%; \cref{app:subsets}). \Cref{fig:shares} shows which candidates it moves traffic between.

\paragraph{Three further model families.}
Adding Qwen3-32B, GLM-4.7-Flash and Nemotron-3-Super-120B on LCB, Omni-MATH and a 2,700-problem subset of MMLU-Pro leaves the picture unchanged (\cref{app:families}). Without deepseek-v4-flash, in a pool of both gpt-oss-20b efforts and the three new models, our router still saves 29.5\% [17.3, 37.2], 12.4\% [$-$0.2, 21.7] and 22.7\% [9.9, 33.6] relative to median pricing, so the saving does not depend on deepseek-v4-flash. This pool keeps gpt-oss-20b at low effort, the cheap candidate the saving needs.

\section{Where it fails}\label{sec:when}
Our probes save little on two datasets, for different reasons. Neither failure could have been forecast from simple statistics: models of headroom fitted on how widely lengths and prices vary do not carry over to held-out datasets (leave-one-dataset-out $R^2$ at most .25; \cref{app:headroom}), so headroom has to be measured, which takes each candidate's actual lengths on a labelled sample.

\paragraph{Little headroom: BigCodeBench.}
BigCodeBench tasks are library-usage functions with short, similar solutions; perfect cost knowledge saves only 10.0\% [$-$1.1, 20.1], and no estimator's saving relative to median pricing excludes zero (ours 0.4 [$-$9.5, 9.1]).

\paragraph{Lengths no estimator predicts: CodeContests.}
CodeContests has as much headroom as APPS (35.8\%), but no estimator captures more than about a quarter of it (ours 9.4\% [0.4, 19.8]). Its lengths depend on the problem (.92--.96 of each candidate's length variance is between problems, not between samples) and are shared across candidates (other candidates' lengths explain .67--.88), against our $R^2$ of .35--.52: length is correlated with empirical difficulty, but difficulty seems hard to predict from these prompts. Empirical difficulty, a problem's solve rate across the other candidates, which requires running and grading every model (\cref{sec:mechanism}), explains about half of length here, as on LCB, but adding it to the cost probe's prediction raises $R^2$ by up to .21, against at most .07 on LCB.

\section{What factors drive cost for a given input?}\label{sec:mechanism}
\paragraph{Length depends mostly on the problem.}
Measured on the actual outputs (\cref{app:mechanism}), for every reasoning candidate on every dataset with repeated samples, most of the variance in log length lies between problems rather than between samples of the same problem (.71--.99). Other candidates' lengths on the same problem explain most of a candidate's length (.42--.96, with one exception on BIG-Bench Extra Hard): the same questions are long for every model.

\paragraph{On math and code, length is correlated with empirical difficulty.}
To ask whether hard problems are the long ones, we measure a problem's difficulty as its solve rate across the other candidates. This \emph{empirical difficulty} requires running and grading every model on the problem, so no router has it; we use it only to explain length, not as a way to price queries. On LCB, AIME and CodeContests, empirical difficulty alone explains about half of each candidate's log length (median $R^2$ .45--.57), and on Omni-MATH and APPS about a third (one training set; \cref{app:mechanism}). The cost probe already predicts most of this: adding empirical difficulty to its prediction raises $R^2$ by at most .07 on LCB, Omni-MATH and APPS, though by up to about .2 on AIME and CodeContests. This is why length from predicted success comes within a few points of the cost probe on LCB and Omni-MATH (\cref{tab:pools}). Its success logits carry more than difficulty, though: pricing from empirical difficulty alone would save less than pricing from them (\cref{app:mechanism}).

\paragraph{On knowledge and mixed-format datasets, it is not.}
On MMLU-Pro, SuperGPQA, BIG-Bench Extra Hard and BigCodeBench, empirical difficulty explains less than a fifth of log length. On MMLU-Pro, pricing from empirical difficulty and each question's subject saves under 4\% relative to median pricing, against 29\% for the cost probe (one training set). Here length from predicted success falls well behind the cost probe: by 18.4\% [11.0, 23.5] on MMLU-Pro and by 22.6\% [12.3, 31.9] on BIG-Bench Extra Hard, whose 23 tasks differ in format (\cref{tab:pools,tab:allest}). We added BIG-Bench Extra Hard and SuperGPQA because a cheap check on one sample per problem suggested this contrast; it held on BIG-Bench Extra Hard and was weaker on SuperGPQA.

\paragraph{When the text alone suffices.}
BIG-Bench Extra Hard has 23 tasks, each with its own fixed instructions and answer format, and they differ widely in length. The text identifies the task, and with it much of the length, so with the full training set text-only estimators predict length as well as the cost probe: MixLLM-style embeddings save 24.4\% relative to median pricing, and ours $-$1.3\% [$-$5.1, 2.7] relative to them. Our probe still needs no second encoder, and with fewer than 100 labelled problems it leads them on average, though within noise (\cref{app:transfer}). On SuperGPQA, the other dataset of mixed subjects, the text estimators fall behind (ours saves 15.7\% [0.1, 27.5] relative to them).

\paragraph{What the probe still misses.}
Our probe predicts length better than every other estimator we tried on every candidate of every dataset, with one exception (\cref{tab:bracket}). The SCX-style estimator, which predicts one length per query from the same activations, comes within .04 of our $R^2$ on some candidates. It still saves 8.8--21.1 points less on LCB, Omni-MATH, MMLU-Pro and AIME, because it fixes the ratio of lengths between candidates for every query (\cref{tab:allest}). Our probe is well below the $R^2$ a perfect predictor of each problem's expected length would reach, the between-problem share in \cref{tab:bracket}. It comes closest on LCB, Omni-MATH and BIG-Bench Extra Hard (on LCB, $R^2$ .71--.75 against .98--.99) and falls furthest below it on the knowledge datasets, above all for deepseek-v4-flash (on MMLU-Pro, .26 against .92). It predicts differences between candidates worst (\cref{app:mechanism}).

\section{How cheaply can a router get a cost prediction?}\label{sec:general}
A deployed router has few labels, adds models, meets benchmarks it was not trained on and serves mixed traffic; we test the cost estimators in all four settings (protocols in \cref{app:settings}) and then ask how small the scout can be.

\paragraph{Few labels.}
\Cref{tab:labels} refits every cost estimator on 20 random training problems and compares it with median pricing fitted on the same problems. Our probe keeps 51--81\% of its full-data saving on each of the five datasets, while the external estimators save at most 10.3\%; at 100 problems ours saves 17.8\% on MMLU-Pro against at most 6.3\% for any of them. With more training problems the external estimators improve but do not catch up: on LCB, Omni-MATH, MMLU-Pro and AIME every one of them still saves significantly less than our probe at the full training set, while ours is within 4.0--6.7 points of its full-data saving at 50 problems on every dataset except MMLU-Pro (11.6). The exception is BIG-Bench Extra Hard, where the task format is visible in the text and the text estimators catch up with more data. Length from predicted success matches ours on LCB and Omni-MATH from 20 problems on, but stays below 7\% at every $n$ on MMLU-Pro and BIG-Bench Extra Hard (\cref{fig:labels}).

\begin{table}[t]
\centering
\footnotesize
\setlength{\tabcolsep}{4pt}
\begin{tabular}{lrrrrr}
\toprule
Cost estimator, 20 training problems & LCB & Omni-MATH & MMLU-Pro & SuperGPQA & BBEH \\
\midrule
Ours (cost probe) & 22.3$^\dagger$ & 22.5$^\dagger$ & 11.9 & 19.5 & 14.2 \\
\midrule
Prompt-feature GBM & $-$1.3 & $-$2.4 & 2.2 & 8.8 & 2.9 \\
MixLLM-style embeddings & $-$3.1 & 2.0 & $-$0.6 & 5.9 & $-$3.2 \\
CARROT-style $k$NN & 1.4 & 3.8 & 2.3 & 10.3 & 6.2 \\
ZeroRouter pricing & 1.2 & 5.6 & 0.1 & $-$1.9 & 1.1 \\
\midrule
\emph{Our ablation: length from predicted success} & 22.3 & 26.7 & 3.2 & 7.0 & $-$10.0 \\
\emph{Ours, fitted on all training problems} & 27.8 & 31.5 & 23.4 & 33.0 & 23.5 \\
\bottomrule
\end{tabular}
\caption{\textbf{Few labels.} Cost saved (\%) relative to median pricing fitted on the same problems, when each cost estimator is fitted on 20 random training problems, two subsets per training set. $^\dagger$: ours exceeds every external estimator (95\% intervals exclude zero). Below the line, for reference: our ablation, which exceeds ours on Omni-MATH, and ours fitted on all training problems, relative to median pricing fitted on all of them.}
\label{tab:labels}
\end{table}

\paragraph{New models.}
To add a model, a router needs its success and cost probes. We hold out one model at a time, with all of its efforts, and fit its probes from $k$ labelled problems while every other candidate keeps its full probes; ours fits two parameters per probe, an offset to the other candidates' predicted log length and a logistic map from their mean success logit. From 10 examples our router saves 19.2--26.7\% on four datasets and 9.7\% on BIG-Bench Extra Hard, against 3.4--17.4\% for ZeroRouter's onboarding, its main use case~\citep{zerorouter}, and leads it on every dataset on average (by 5.2--13.0 points), though at 10 examples no interval excludes zero; at 50 examples ours leads significantly on LCB and SuperGPQA (\cref{tab:onboard}). The gain concentrates on deepseek-v4-flash, the one model whose family is not otherwise in the pool.

\paragraph{New benchmarks.}
We then move a whole router to a benchmark it has not seen, training both probes on the other datasets of the same domain (code, or math and knowledge) with no target labels (\cref{tab:transfer}). On LCB and AIME the transferred router keeps 94--96\% of its in-domain saving, and on MMLU-Pro and APPS it exceeds it, leading the best transferred external estimator by 7.5--28.9 points (significantly on LCB, MMLU-Pro and AIME). The prefill router's rule, the same transferred success probes with median pricing, saves at most 9.0\% on these four and loses money on LCB, so the transferred cost probe carries the result. Transfer is weaker on SuperGPQA, Omni-MATH and BIG-Bench Extra Hard, where a few target labels help. Length from predicted success transfers to fewer benchmarks: the map from difficulty to length differs between benchmarks, while the map from the scout's activations to length carries over to more of them (\cref{app:transfer}).

\begin{table}[t]
\centering
\footnotesize
\setlength{\tabcolsep}{4pt}
\begin{tabular}{lrrrr}
\toprule
 & Ours, & Ours, & Prefill-router rule, & Best external, \\
Sources $\to$ target & in-domain & transferred & transferred & transferred \\
\midrule
APPS, BCB, CC $\to$ LCB & 27.8 & 26.0 & $-$9.4 & $-$3.0 (ZeroRouter) \\
LCB, BCB, CC $\to$ APPS & 16.8 & 31.5 & 9.0 & 23.7 (ZeroRouter) \\
Omni, AIME, SGPQA, BBEH $\to$ MMLU-Pro & 23.4 & 26.6 & 1.1 & 18.1 ($k$NN) \\
Omni, MMLU-Pro, SGPQA, BBEH $\to$ AIME & 14.9 & 14.3 & $-$1.7 & 6.8 ($k$NN) \\
Omni, MMLU-Pro, AIME, BBEH $\to$ SuperGPQA & 33.0 & 22.5 & $-$11.2 & 17.6 (MixLLM) \\
Omni, MMLU-Pro, AIME, SGPQA $\to$ BBEH & 23.5 & $-$3.4 & $-$21.0 & $-$6.5 (ZeroRouter) \\
MMLU-Pro, AIME, SGPQA, BBEH $\to$ Omni-MATH & 17.8 & 4.0 & $-$23.4 & 1.7 (MixLLM) \\
\bottomrule
\end{tabular}
\caption{\textbf{A router moved to a new benchmark.} Cost saved (\%) relative to the target's own median pricing. Ours: success and cost probes trained on the sources, with every choice made on their calibration splits; external estimators transfer the same way, each with its own success model. Prefill-router rule: the transferred success probes with median pricing. Ours is significantly ahead of every alternative in every training set on LCB and MMLU-Pro (9 of 10 on AIME). SGPQA: SuperGPQA; Omni: Omni-MATH's 500-problem set.}
\label{tab:transfer}
\end{table}

\paragraph{Mixed traffic.}
Much of what quality routers gain can be had by knowing a query's dataset or task type~\citep{llmrouterbench,tasktype}; a cost router might likewise get most of its saving by segmenting traffic. We route one mixture of all nine benchmarks~\citep{routerbench,llmrouterbench} and compare with routers given each query's true benchmark (\cref{tab:mixed}), an optimistic baseline, since a deployed router would have to infer it with some error. Pricing each candidate at its median length within the query's benchmark saves 12.2\% over one median per candidate; our single probe, not given the benchmark, saves 31.8\%, or 22.4\% [16.5, 27.6] beyond benchmark identity, with intervals excluding zero in all ten training sets. MixLLM, whose encoder is fine-tuned to separate the benchmarks during training but uses only the query at test time, saves 8.8\% beyond benchmark identity; ours saves 14.9\% [9.9, 20.0] beyond MixLLM. To test whether the probe captures more than benchmark and difficulty, we fit length from predicted success within each benchmark, given the true benchmark: it saves 24.4\%, and the cost probe still saves 9.9\% [5.2, 14.2] beyond it.

\begin{table}[t]
\centering
\footnotesize
\setlength{\tabcolsep}{4pt}
\begin{tabular}{lrrr}
\toprule
 & \multicolumn{2}{c}{Cost saved relative to} & \\
\cmidrule(lr){2-3}
Router prices each candidate by & one median & median per benchmark & Ours vs.\ it \\
\midrule
\emph{Baselines} & & & \\
One median length (the prefill router's rule) & -- & $-$13.9 & 31.8 [26.1, 36.7] \\
Median length per benchmark & 12.2 & -- & 22.4 [16.5, 27.6] \\
Mean length per benchmark & 13.5 & 1.5 & 21.2 [15.9, 26.0] \\
MixLLM-style embeddings & 18.9 & 7.6 & 16.0 [11.1, 19.9] \\
MixLLM (benchmarks as domains) & 19.9 & 8.8 & 14.9 [9.9, 20.0] \\
\midrule
\emph{Our ablation: length from predicted success} & & & \\
Fitted per benchmark & 24.4 & 13.8 & 9.9 [5.2, 14.2] \\
Fitted on the mixture & 11.7 & $-$0.5 & 22.8 [17.8, 27.8] \\
\midrule
\textbf{Ours} (one probe, benchmark not given) & \textbf{31.8} & \textbf{22.4} & -- \\
\midrule
\emph{Actual cost (headroom)} & 46.4 & 38.9 & \\
\bottomrule
\end{tabular}
\caption{\textbf{One router for mixed traffic.} All nine benchmarks in one mixture (150 test problems each), with the same five candidates and success probes in every row. Cost saved (\%) at matched accuracy, with 95\% intervals resampled within each benchmark; every ``ours vs.\ it'' interval excludes zero in all ten training sets. Estimators with access to benchmark identity are given each query's true benchmark (an optimistic baseline); ours is not (\cref{app:baselines}).}
\label{tab:mixed}
\end{table}

\paragraph{Scout size.}
Savings do not grow with the scout's size (\cref{tab:scale}): from 1.7B to 14B, every Qwen3 scout saves 24.9--28.8\% on LCB, 27.9--30.7\% on Omni-MATH and 19.4--20.7\% on MMLU-Pro relative to median pricing with its own success probes, while its forward pass costs about seven times more at 14B than at 0.6B. Counting each router's own pass, ours (4B) spends significantly less than the 14B router on LCB and MMLU-Pro (by 7.3\% [2.2, 11.2] and 10.5\% [2.3, 15.5]) and than the 8B router on MMLU-Pro (6.2\% [1.1, 12.4]). A 0.6B scout predicts length worse (ours is ahead by 7.3--8.2 points before its cheaper pass is counted), and a 1.7B scout does about as well as ours. Scouts from three other families save 16.9--21.4\% on MMLU-Pro and 25.1--28.2\% on Omni-MATH (\cref{tab:size}).

\begin{table}[t]
\centering
\footnotesize
\setlength{\tabcolsep}{3.5pt}
\begin{tabular}{lrrrrrrrrr}
\toprule
 & \multicolumn{3}{c}{LiveCodeBench} & \multicolumn{3}{c}{Omni-MATH} & \multicolumn{3}{c}{MMLU-Pro} \\
\cmidrule(lr){2-4}\cmidrule(lr){5-7}\cmidrule(lr){8-10}
Scout & vs.\ median & \$/1k & Ours vs.\ it & vs.\ median & \$/1k & Ours vs.\ it & vs.\ median & \$/1k & Ours vs.\ it \\
\midrule
Qwen3-0.6B & 24.1 & .005 & 3.5 & 27.9 & .002 & 6.5 & 17.0 & .002 & 4.8 \\
Qwen3-1.7B & 27.6 & .007 & $-$1.6 & 29.0 & .003 & 3.4 & 19.4 & .004 & 2.5 \\
Qwen3-4B & 25.5 & .015 & 2.7 & 28.1 & .006 & 3.5 & 20.6 & .007 & 3.2 \\
Qwen3-8B & 24.9 & .022 & 3.6 & 30.7 & .008 & 0.6 & 20.7 & .011 & \textbf{6.2} \\
Qwen3-14B & 28.8 & .036 & \textbf{7.3} & 27.9 & .014 & 9.8 & 20.1 & .018 & \textbf{10.5} \\
\midrule
Qwen3-4B-2507 (ours) & 27.8 & .015 & -- & 31.5 & .006 & -- & 23.4 & .007 & -- \\
\bottomrule
\end{tabular}
\caption{\textbf{Scout size.} Qwen3 scouts from 0.6B to 14B (the original releases, which switch between thinking and non-thinking modes) and ours (4B Instruct-2507), each feeding both probes. vs.\ median: cost saved (\%) relative to median pricing with the row's own success probes. \$/1k: the scout's forward pass per 1,000 queries on one H100 at \$2 per hour. Ours vs.\ it: cost saved (\%) by our router relative to the row's, each paying for its own scout's pass, at our spend in the middle of the accuracy band (one training set); bold: 95\% interval excludes zero. Scouts from other model families in \cref{tab:size}.}
\label{tab:scale}
\end{table}

\section{Controls}\label{sec:ablations}
\paragraph{A text-embedding model of the same size.}
Reading cost from Qwen3-Embedding-4B~\citep{qwen3emb}, a text-embedding model of the scout's size, with any of three probes and our success predictions, or building a fully text-based router on it, saves less than our router: by 9.7--23.1\% (embedding cost probes) and 6.8--18.5\% (text router) on LCB, Omni-MATH, MMLU-Pro and AIME, with intervals excluding zero, and on average on the other five datasets, where it is never significantly ahead (\cref{tab:textemb}).

\paragraph{Is fine-tuning needed?}
Fully fine-tuning DistilBERT 66M, jina-embeddings 137M or Qwen3-0.6B on the same length labels, as CARROT does with RoBERTa~\citep{carrot}, saves less than our probe on Omni-MATH and MMLU-Pro (by 11.9--25.6\%, one training set) and on LCB (by 6.5\% [$-$1.2, 12.2] for Qwen3-0.6B); the same Qwen3-0.6B, frozen, with a linear probe on its activations, beats its fine-tuned copy, consistent with full fine-tuning distorting good pretrained features~\citep{kumar2022finetuning} (\cref{app:size}).

\paragraph{Is the prefill router's success pipeline needed?}
At our training sizes (a few hundred problems per benchmark, far fewer than the prefill router was trained on), our logistic probes on fixed layers tie a reimplementation of its success pipeline~\citep{prefill} (per-target layer, pooling and PCA selection, then an MLP ensemble) on four of five datasets and trail it only on SuperGPQA, where neither predicts success well (one training set; \cref{app:size}). The same-size text embedding predicts success worse on every dataset.

\section{Deployment}\label{sec:deploy}
\paragraph{Routing versus single models.}
A random mix of candidates with fixed probabilities reaches any point on the convex hull of the candidates' cost--accuracy points without a router; it is the baseline a router has to beat~\citep{routerbench}. At matched accuracy our router spends 22.9--39.2\% less than the mix on LCB, Omni-MATH and MMLU-Pro and 14.9--46.6\% less on AIME, SuperGPQA and BIG-Bench Extra Hard, and ties it on the other three datasets. No other estimator beats the mix on those three either, and on BigCodeBench neither does perfect cost knowledge. With median pricing the same success probes do not beat the mix on MMLU-Pro or AIME: there the cost estimate determines whether routing pays at all.

At the top of the frontier the cost estimate stops mattering: the most accurate policy sends each query to the candidate with the highest predicted success, so it is the same for every cost estimator. deepseek-v4-flash is the most accurate candidate or within 1.6 points of it on every dataset, and the router's most accurate policy ends 0.2--2.7 points below the most accurate candidate on seven datasets; large-scale evaluations likewise find routers close to assigning each dataset its best single model~\citep{llmrouterbench}. Larger success predictors are unlikely to pay for themselves here: our 4B pass already costs 2--30\% of the saving and savings grow little from a 0.6B to a 4B scout (\cref{app:deploy}). Below the top, giving up five points of accuracy saves 14.9--40.3\% relative to deepseek-v4-flash alone on six datasets, 1.5--1.8$\times$ what median pricing saves. Under a hard cap on each call's cost, the router beats deepseek-v4-flash alone at the tightest cap we tested (by 8.2 and 7.0 points of accuracy on LCB and MMLU-Pro) and ties or trails it by up to 3 points at looser caps (\cref{app:deploy}).

\paragraph{Deployable policies and the router's own cost.}
A deployed router must fix $V$ in advance. Choosing, for each of ten accuracy targets, the cheapest $V$ that reaches it on calibration, our deployed policies save 19.6--37.9\% relative to median pricing's on six datasets, cost 1.0--3.3\% more than our own test frontier and miss their target by 2.2--5.3 points (\cref{tab:deploy}). The features come from one 4B forward pass over the prompt (19--28\,ms on one H100), whose cost a prefill router that predicts success already incurs; the cost probe adds nothing to it (\cref{app:deploy}).

\section{Discussion and Limitations}
\paragraph{When does per-query cost pay?}
For reasoning candidates perfect cost knowledge is worth 29--61\% on eight of nine datasets, and much less for non-reasoning candidates on the same problems; how much it is worth on a new dataset is not predictable from how widely lengths and prices vary, but it is cheap to measure. The contrast is consistent with CARROT, which found that per-query cost adds little over each model's average cost on RouterBench, whose eleven models include no reasoning model~\citep{routerbench}, and on SPROUT, where one of fourteen is (o3-mini)~\citep{carrot}. The saving comes from deciding, query by query, whether a strong candidate whose cost varies widely is worth calling instead of a cheap one. For reasoning candidates the variation depends mostly on the problem, so the estimator is what limits capture. Where length is correlated with empirical difficulty, even length from predicted success captures most of it. Where it is not, a predictor trained on length is needed, and on MMLU-Pro only the cost probe captures it. Capture there is still low (.40), which leaves room for better cost predictors. One-shot software-engineering patches behave like the coding datasets (\cref{app:agents}).

\paragraph{Generalisation.}
In-domain and with full training data, a router could take length from predicted success and lose little on most datasets. The case for a dedicated cost probe is that it holds up where a deployed router actually operates: with few labels, with new models, on new benchmarks and in mixed traffic, where the text-based estimators and ZeroRouter fall further behind it and length from predicted success holds up only on benchmarks like those it was trained on (\cref{sec:general}). A benchmark unlike its sources still needs some labels: on Omni-MATH and BIG-Bench Extra Hard the transferred router saves little.

\paragraph{Limitations.}
Each candidate needs output-length labels, though a few dozen suffice for most datasets. Exceeding the most accurate single candidate would take better success predictions than ours; the cost probe does not change the top of the frontier. At these model prices a larger success predictor would cost a large share of the saving it is meant to improve, but the case for one grows with the price of the routed models. Telling apart two strong candidates of similar price is the open part of the problem: billing gpt-oss-120b at high effort at deepseek-v4-flash's token rates, perfect per-query cost knowledge would save 11--36\% within that pair, but our router saves at most 7.0\% [2.4, 14.0] (Omni-MATH) relative to calling deepseek-v4-flash alone at its accuracy. The saving also needs a cheap candidate that solves a good share of problems. The main experiments use five candidates from two model families, with three further families on three datasets, and transfer is evaluated across benchmarks that share the routed models. Each interval covers one comparison, with no correction for the number of comparisons, and treats test problems as independent, though problems from one contest or subject may not be. A few analyses use one training set. The Omni-MATH and MMLU-Pro test sets have one sample per problem and candidate. Of the five main candidates only deepseek-v4-flash is pinned to a provider; billed rates come from one collection, and prices change within days. The ZeroRouter, MixLLM and CARROT baselines are reimplementations.

\section{Conclusion}
A small frozen scout's prefill activations, already used to predict success, also predict cost. Per-query cost is worth predicting for reasoning models, whose output lengths vary widely within a model, and much less for non-reasoning ones on the same problems. On held-out test problems at billed prices, the cost probe reduces spending at matched accuracy on eight of nine datasets, and it is ahead in 69 of 72 comparisons with external estimators, significantly in 50; length from predicted success matches it only where output length is correlated with empirical difficulty, and only in domain. With few labels and for new models the cost probe keeps most of its saving, and on four of seven new benchmarks it does so without any target labels; in mixed traffic it saves a fifth of the cost beyond pricing by each query's benchmark.

\section*{Reproducibility}
Every number in the paper is computed from stored per-call records and prefill activations, with every split fixed by a seed. Prices and provider behaviour change over time, so a new collection would not reproduce billed costs exactly. We plan to release the code.

\section*{Acknowledgements}
This research was funded in part by Mitacs through the Mitacs Accelerate program and from the direct support of ServiceNow. Compute was provided by ServiceNow Research. We used Claude (Anthropic) to write the analysis and experiment code and to help draft the article text.

\bibliography{references}
\bibliographystyle{tmlr}

\appendix
\FloatBarrier
\section{Pools}\label{app:pools}
\begin{table}[h]
\centering
\footnotesize
\setlength{\tabcolsep}{4pt}
\resizebox{\linewidth}{!}{%
\begin{tabular}{lrccrrr}
\toprule
 & Train / calib. / test & Samples per candidate & Samples per candidate & Test acc. & Median output & 90th/10th \\
Dataset & problems & (train) & (test) & (\%) & per candidate & pct.\ ratio \\
\midrule
LiveCodeBench~\citep{lcb} & 441 / 110 / 341 & 16/12/8/6/3 & 16/12/8/6/3 & 52--90 & 0.7--7.0k & 11--121 \\
Omni-MATH~\citep{omni} & 275 / 75 / 1,000 & 4/3/3/2/2 & 1/1/1/1/1 & 42--71 & 1.4--10.5k & 10--126 \\
MMLU-Pro~\citep{mmlupro} & 550 / 150 / 6,500 & 4/3/3/2/2 & 1/1/1/1/1 & 62--86 & 0.2--1.6k & 7--29 \\
AIME 1983--2024~\citep{aime} & 513 / 139 / 281 & 4/3/3/2/2 & 4/3/3/2/2 & 57--98 & 1.8--7.7k & 4--16 \\
APPS~\citep{apps} & 489 / 123 / 377 & 1/1/1/1/1 & 1/1/1/1/1 & 58--86 & 0.6--10.3k & 10--91 \\
BigCodeBench~\citep{bigcodebench} & 544 / 136 / 421 & 16/12/8/6/3 & 16/12/8/6/3 & 45--53 & 0.4--2.1k & 2--4 \\
CodeContests~\citep{codecontests} & 346 / 86 / 266 & 4/3/3/2/2 & 4/3/3/2/2 & 49--86 & 0.8--13.1k & 5--37 \\
SuperGPQA~\citep{supergpqa} & 550 / 150 / 300 & 4/3/3/2/2 & 4/3/3/2/2 & 31--68 & 0.2--3.3k & 7--25 \\
BIG-Bench Extra Hard~\citep{bbeh} & 550 / 150 / 300 & 4/3/3/2/2 & 4/3/3/2/2 & 23--65 & 0.4--16.4k & 14--26 \\
\bottomrule
\end{tabular}}
\caption{\textbf{Pools.} Split sizes (problems with a valid answer from every candidate), median samples per candidate (in \cref{tab:routes}'s order), and ranges over candidates of test accuracy, median output length (tokens) and the 90th/10th-percentile length ratio. Omni-MATH and MMLU-Pro train on Omni-MATH-500 and a 1,000-question subject-stratified sample and test on 1,000 and 6,500 further problems; the other datasets use a random held-out split. Billed rates (USD per million input / output tokens): gpt-oss-20b 0.029 / 0.134, deepseek-v4-flash (StreamLake) 0.042 / 0.084, gpt-oss-120b 0.020 / 0.257.}
\label{tab:poolstats}
\end{table}

\FloatBarrier
\section{Implementation details}\label{app:protocol}
\subsection{Data, candidates and labels}
All calls go through OpenRouter. deepseek-v4-flash is pinned to StreamLake by restricting OpenRouter's provider routing to that provider with fallbacks disabled; the gpt-oss candidates are served by whichever provider OpenRouter selects. LiveCodeBench has 892 problems (contests from September 2023 to April 2025), split at random into 441 training, 110 calibration and 341 test problems. Omni-MATH and MMLU-Pro have 275/75 and 550/150 training/calibration problems, from the 500-problem Omni-MATH set and a 1,000-question subject-stratified sample of MMLU-Pro, and 1,000 and 6,500 disjoint test problems. The six further datasets are split at random (\cref{tab:poolstats}). Training problems have 2--16 samples per candidate; Omni-MATH and MMLU-Pro test problems have one. Every sample is scored by the benchmark's own grader (code execution against the tests for the coding datasets; answer matching for the others), and evaluated problems need a valid answer from every candidate.

\subsection{Prices}
Effective rates are least-squares fits of each call's billed cost on its input and output tokens, per model (\cref{tab:poolstats}). For the gpt-oss models they average over the providers OpenRouter chose, which billed gpt-oss-120b from 0.17 to 0.60 USD per million output tokens. A problem's actual cost on a candidate is the mean billed cost of its valid samples on the Omni-MATH and MMLU-Pro test sets, and the mean of its valid samples' input and output tokens at the effective rates elsewhere. Two checks over ten training sets: fitting the rates on training calls only (the same for deepseek-v4-flash and gpt-oss-120b, a 15\% higher output price for gpt-oss-20b), and costing the Omni-MATH and MMLU-Pro test sets as tokens at the effective rates. Each moves our saving relative to median pricing by at most 1.0 point, and no comparison in \cref{tab:main} flips sign or loses significance. SuperGPQA and BIG-Bench Extra Hard were collected later, when deepseek-v4-flash was billed at about twice its earlier rate; we price them like every other dataset (at their own billed rates our saving there would be 4.9 and 1.1 points higher, on one training set).

\subsection{Probes}
The features are the mean- and last-token activations of eight layers of Qwen3-4B (Instruct-2507; Thinking-2507 on Omni-MATH), concatenated and shifted and scaled to zero mean and unit variance on the training problems. Success probes are per-candidate L2 logistic regressions, with the regularization chosen on calibration AUC from nine values and Platt scaling fitted on calibration. Cost probes are per-candidate ridge regressions of log mean output length, with the regularization chosen by cross-validation within the training problems; the exponentiated prediction is corrected by Duan's smearing from the training residuals and rescaled to the candidate's training mean.

\subsection{Evaluation}
\paragraph{Frontiers and matched accuracy.} Each value $V$ on a geometric grid of 300 values gives a deterministic policy (\cref{eq:route}) and a test point (mean actual cost, mean accuracy). An estimator's frontier is the upper concave hull of its points, and its cost at a target accuracy is linearly interpolated on the hull. Two estimators are compared at 12 accuracies evenly spaced over the interior 90\% of the accuracy range both frontiers reach; the reported saving is one minus the geometric mean of their cost ratios. Which problems are used for training matters at these sizes, so every result is a mean over ten training sets. A training set reshuffles the training and calibration problems (with a fixed seed per training set) into splits of the same sizes and keeps the test problems; the success probes and every estimator are refitted and every configuration is re-chosen on its calibration problems. Within a training set, intervals resample test problems with replacement (300 paired resamples, the same for both estimators and for every training set, all candidates of a problem kept together) with that set's probes fixed; reported intervals are percentiles of the 3,000 resamples pooled over the ten training sets, so they reflect both the choice of training problems and the test problems, and reported values are means of the ten point estimates.
\paragraph{Analyses on one training set.} Every table and every number in the main text is a mean over the ten training sets unless marked ``one training set''. Those marked use one fixed split of each dataset and compare estimators within it: held-out samples, rescaled prices, the length drivers of \cref{sec:mechanism}, fine-tuned encoders on Omni-MATH and MMLU-Pro, parts of the transfer analysis and the prefill router's success pipeline.

\subsection{Reimplemented estimators}\label{app:baselines}
Every estimator predicts each candidate's log mean output length from information available before generation, is fitted on the training problems only, and is post-processed like our probe (residual smearing, rescaled to the candidate's training mean); configurations are chosen on the calibration split.

\emph{Length from predicted success}, our ablation, fits each candidate's log length by ridge regression on the success logits of all five candidates and their squares, so it uses the scout's activations only through the success probes.

\emph{Prompt-feature GBM} follows latency-aware routing~\citep{latencyrouting}: histogram gradient boosting (300 iterations, learning rate 0.05, minimum leaf 10) on 44 prompt statistics (length, line count, count and magnitude of numbers, math delimiters, examples, constraint markers, code fences, inequality signs and 34 keyword counts), with scikit-learn's histogram boosting in place of LightGBM.

\emph{MixLLM-style}~\citep{mixllm}: a jina-embeddings-v2 137M embedding~\citep{jina} of the prompt (the code variant on LiveCodeBench), fed to the average of an MLP, a random forest and a $k$-nearest-neighbour regressor per candidate. It keeps MixLLM's length regressors but neither its encoder nor its loss.

\emph{MixLLM}~\citep{mixllm} as published, in mixed traffic only: bert-base-uncased, mean-pooled, fine-tuned with the paper's domain loss (each embedding pulled toward its domain's centre, centres pushed apart), with the nine benchmarks as domains in place of InsTag tags grouped by hand; the paper's tags need a 13B tagger and give sub-topics our benchmarks do not label, and within one benchmark the loss has nothing to separate, so the per-dataset tables use the MixLLM-style version. The paper gives no normalization or optimizer settings, so embeddings and learnable centres are normalized with temperature 0.1 (AdamW, learning rate $2\cdot10^{-5}$, batch 32, up to three epochs, the epoch chosen on calibration log-length error). Each candidate's length regressor is the MLP, random forest or $k$-nearest-neighbour regressor with the lowest calibration error (the random forest in 48 of 50 cases).

\emph{CARROT-style $k$NN}~\citep{carrot}: the authors' local variant, MiniLM-L12~\citep{minilm} sentence embeddings with uniform cosine $k$NN, $k$ chosen by five-fold cross-validation on training problems.

\emph{Route-To-Reason}~\citep{routetoreason}, as published: a frozen all-mpnet-base-v2 embedding of the query, concatenated with a fixed embedding of a short text description and a learned 64-dimensional embedding of the model and of its effort (the paper's reasoning strategy), fed to one MLP shared by all candidates (hidden size 768) trained with squared error on output length (Adam, learning rate $10^{-3}$, batch 32, up to 100 epochs, early stopping on calibration loss); the same network on log length saves within 5 points of it on every dataset.

\emph{One length per query (SCX-style)}~\citep{scx}: SCX Router predicts one expected output length per task rather than per model; we predict one per query from our scout's activations (ridge onto the candidate-centred mean log length) and add each candidate's training mean.

\emph{ZeroRouter}~\citep{zerorouter}: stage 1 fits a $D$-dimensional two-parameter item-response model to the five candidates' training outcomes (MAP with Gaussian priors; the paper uses stochastic variational inference over about 200 leaderboard models). Stage 2 places a new query in this latent space by regressing its item parameters $(\log\alpha, b)$ on the query. \emph{As published}, stage 2 fine-tunes DistilBERT~\citep{distilbert} with 11 linguistic features of the prompt (40 epochs, batch 32, learning rate $3\cdot10^{-5}$, squared error on the standardized parameters, the epoch chosen on a held-out tenth of the training problems; the paper does not list its features, so ours are length, word and sentence statistics, two readability scores, the shares of digits and math symbols, and bracket depth). \emph{With our 4B scout}, stage 2 is a ridge regression on our scout's features after PCA to 256 dimensions. Both use the paper's success model $\sigma(\alpha^\top(\theta_m-b))$ and its pricing, the candidate's mean training length within $K$ quantile bins of $s=\alpha^\top b$, with $D\in\{1,5\}$ and $K\in\{5,10,20\}$ chosen on calibration.

\Cref{tab:allest,tab:main} use the 4B-scout version, from the same run as the rest of the table. The version as published was fitted in a separate run on the ten training sets, which also refits the fine-tuned encoder, so it varies more between runs. With DistilBERT, ours saves 27.4, 23.9, 23.6, 14.2, 14.5, 3.9, 7.9, 35.5 and 17.5\% relative to ZeroRouter's pricing with our success, and 23.5, 19.6, 19.5, 11.6, 7.1, $-$10.6, 3.3, 35.9 and 28.6\% relative to it with its own success (datasets in the order of \cref{tab:allest}). Our scout's features predict success better on every dataset (AUC .55--.81 against .52--.73). Elsewhere in the paper ZeroRouter means the 4B-scout version, the stronger of the two. With five candidates a higher-dimensional latent does not improve its success predictions, and on MMLU-Pro adding up to 53 Open LLM Leaderboard models to stage 1 did not either.

The fine-tuned encoders of \cref{sec:ablations} are mean-pooled with a linear head on per-candidate standardised log length, trained with masked squared error (AdamW, 6\% warm-up then linear decay, maximum 512 tokens for DistilBERT and 2,048 for the others); learning rate (two values per encoder) and epoch (up to 10, or 20 when the best epoch was the last) are chosen on calibration error, and smearing and rescaling use the calibration problems.

\FloatBarrier
\section{Headroom}\label{app:headroom}
\paragraph{The headroom is a practical upper bound.}
The bound is not strict, because the success predictions stay imperfect, but every estimator stays below it on all nine pools. Pricing each problem at its actual cost assumes that the random variation of the samples it is scored on is known in advance, which no predictor of expected cost can foresee. Where test problems have several samples per candidate, we split each problem's samples at random into a half that prices and a half that is scored (20 splits). On one training set, pricing at the held-out half's mean length, a target an expected-cost predictor could reach, saves 41.2\% on LiveCodeBench against 45.9\% for perfect cost knowledge on the same halves; the gap is larger where samples are few (AIME 20.1 against 30.1, Omni-MATH-500 23.6 against 35.9, CodeContests 23.4 against 38.9) and small on SuperGPQA (52.7 against 57.1). With one or two samples per half the held-out mean is itself noisy, so the achievable headroom lies between the two; on LiveCodeBench, where it is tightest, the cost probe captures .81 of the held-out bound against .72 of the actual one. The capture figures in \cref{tab:pools} are therefore conservative.

\paragraph{Headroom cannot be forecast from simple statistics.}
Before fitting, we chose four linear models that predict a pool's headroom from statistics that need no predictor: (A) how widely prices differ between candidates (s.d.\ over candidates of log mean cost per call), (B) how much log length varies in its shared and candidate-specific parts, (C) both, and (D) the log ratio of that variation to the squared price spread. Fitted over the 138 candidate subsets of \cref{app:subsets} (full pools, leave-one-out subsets and pairs), they explain .32, .18, .37 and .11 of the variance in headroom, but on held-out datasets only .25, $-$.23, .01 and $-$.01. Headroom depends on which candidates solve which problems, not only on how lengths and prices vary.

\FloatBarrier
\section{Non-reasoning candidates}\label{app:nonreason}
\paragraph{Setup.} deepseek-v4-flash with thinking disabled uses the same provider and sampling as with thinking; the four other non-reasoning models are each pinned to one provider. Every non-reasoning candidate has one sample per problem, with output capped at 16k tokens. Qwen3-235B-A22B-Instruct has training answers for 241 of 275 Omni-MATH and 485 of 550 MMLU-Pro problems (rate limits).
\begin{table}[h]
\centering
\footnotesize
\setlength{\tabcolsep}{4pt}
\begin{tabular}{lrrrrrr}
\toprule
 & \multicolumn{2}{c}{Headroom} & \multicolumn{2}{c}{Ours vs.\ median} & \multicolumn{2}{c}{Capture} \\
\cmidrule(lr){2-3}\cmidrule(lr){4-5}\cmidrule(lr){6-7}
Test set & reasoning & non-reas. & reasoning & non-reas. & reasoning & non-reas. \\
\midrule
LiveCodeBench (341) & 40.1 & 21.4 & 27.8 & 12.1 (5.4) & .69 & .53 \\
Omni-MATH (1,000) & 53.1 & 40.5 & 31.5 & 6.8 (1.6) & .59 & .17 \\
MMLU-Pro (2,000) & 59.8 & 26.0 & 25.8 & 0.8 (2.2) & .43 & .03 \\
\bottomrule
\end{tabular}
\caption{\textbf{Reasoning versus non-reasoning candidates on the same problems.} Headroom, our saving relative to median pricing (s.d.\ over training sets) and capture, for five reasoning and five non-reasoning candidates, on the test problems every candidate of both pools answers validly (MMLU-Pro: a 2,000-problem subset). Here perfect cost knowledge prices each problem at its billed cost, so Omni-MATH's reasoning headroom differs slightly from \cref{tab:pools}. The non-reasoning saving excludes zero in every training set on LCB and Omni-MATH and in none on MMLU-Pro.}
\label{tab:nonreason}
\end{table}
Adding the non-reasoning candidates to the reasoning pool lowers our saving (14.8\% on Omni-MATH, 7.6\% on MMLU-Pro, 19.2\% on LCB). Non-reasoning lengths do vary (s.d.\ of log length .53--1.73 per candidate), but the variation has two sources that the prompt predicts poorly. Llama-3.1-8B, at its default sampling settings, falls into repetition until the 16k cap on 9--12\% of math problems (four in five capped answers loop), so its mean output is 4--6$\times$ its median and its log-length $R^2$ is below zero on every set. The loops come from the model, not the provider: redrawn with another provider, it loops about as often. Qwen3-30B-A3B-Instruct, asked for code only on LCB, reasons inside code comments on hard problems until the cap (9\% of problems), so its lengths span 142$\times$ between its 10th- and 90th-percentile problems ($R^2$ .27): the largest non-reasoning headroom comes from a non-reasoning model that reasons anyway.

\paragraph{Every estimator on non-reasoning candidates.}
Non-reasoning candidates combine both failures (\cref{tab:nonreason}): less headroom, and on Omni-MATH and MMLU-Pro a capture of .17 and .03. No other estimator does better. Fitting every estimator of \cref{tab:allest,tab:textemb} on the non-reasoning candidates (ten training sets), none of mean length, prompt features, MixLLM-style or nearest-neighbour embeddings, ZeroRouter's pricing and the three Qwen3-Embedding-4B probes saves significantly relative to median pricing on either dataset (at most 1.9\%); on Omni-MATH ours is ahead of each by 4.8--12.2 points with length from predicted success tied, and on MMLU-Pro every estimator, ours included, is within noise of median pricing. Most of their length is nonetheless predictable: on Omni-MATH our probe explains .55--.75 of four candidates' log length, about what the other non-reasoning candidates' lengths on the same problem explain (.48--.70). The exception is Llama-3.1-8B: no estimator predicts its length (best $R^2$ .10--.22 across the three datasets) and neither do other models' lengths (.14--.40). What can be predicted is worth little here, and the most erratic candidate cannot be predicted at all. On LCB the probes capture .53 of a smaller headroom, with length from predicted success and the Qwen3-Embedding-4B ridge probe close behind (11.3\% and 10.4\% against 12.1\%).

\FloatBarrier
\section{Robustness of the main comparison}\label{app:allest}
\Cref{tab:main} gives the main comparison with 95\% intervals on LCB, Omni-MATH and MMLU-Pro, and \cref{tab:resplit} repeats it on ten random re-splits that also reshuffle the test problems.
\begin{table}[h]
\centering
\footnotesize
\setlength{\tabcolsep}{2.5pt}
\resizebox{\linewidth}{!}{%
\begin{tabular}{lrrr}
\toprule
Cost saved by ours vs. & LCB ($n$=341) & Omni-MATH ($n$=1,000) & MMLU-Pro ($n$=6,500) \\
\midrule
Median pricing & 27.8 [13.9, 35.6] & 31.5 [23.9, 37.7] & 23.4 [17.1, 29.0] \\
Mean pricing & 27.8 [19.5, 34.6] & 34.3 [26.3, 40.1] & 25.8 [20.4, 30.6] \\
Prompt-feature GBM & 20.7 [10.1, 29.0] & 25.8 [15.9, 35.0] & 20.0 [13.9, 25.9] \\
MixLLM-style embeddings & 16.2 [7.9, 22.6] & 24.7 [15.9, 31.7] & 15.5 [10.6, 20.5] \\
CARROT-style $k$NN & 17.5 [7.9, 25.4] & 23.2 [14.9, 29.3] & 14.3 [8.3, 20.7] \\
ZeroRouter pricing, our success & 16.8 [9.4, 25.3] & 21.0 [10.5, 31.6] & 22.0 [16.5, 27.9] \\
ZeroRouter pricing, its success & 19.3 [7.7, 28.2] & 14.2 [$-$4.7, 39.3] & 18.4 [5.9, 24.2] \\
\midrule
\emph{Headroom: perfect cost knowledge vs.\ median} & 40.1 [29.0, 46.9] & 51.9 [44.3, 56.8] & 58.0 [53.9, 60.8] \\
\bottomrule
\end{tabular}}
\caption{\textbf{Cost saved at matched accuracy, billed costs.} Cost saved (\%) by our cost probe relative to each estimator, with 95\% intervals. Every row uses our success predictions except ZeroRouter pricing with its own success model; both ZeroRouter rows predict item parameters from our scout's activations. All intervals exclude zero except against ZeroRouter with its own success model on Omni-MATH (ours ahead in 7 of 10 training sets). Last row: the headroom, of which the cost probe captures 69\%, 61\% and 40\%. Estimators in \cref{app:baselines}.}
\label{tab:main}
\end{table}

\begin{table}[h]
\centering
\footnotesize
\setlength{\tabcolsep}{4pt}
\begin{tabular}{lrrrcrc}
\toprule
 & Headroom & Ours vs & Ours $-$ best & Splits ahead & Ours $-$ length & Splits ahead / \\
Dataset & & median & external & of all external & from pred.\ success & behind \\
\midrule
LiveCodeBench & 45.9 $\pm$ 3.0 & 34.1 $\pm$ 3.5 & 12.9 $\pm$ 2.9 & 10 & 2.2 $\pm$ 2.6 & 4 / 0 \\
Omni-MATH & 41.7 $\pm$ 2.5 & 19.0 $\pm$ 3.3 & 12.0 $\pm$ 3.8 & 9 & $-$1.4 $\pm$ 2.6 & 1 / 2 \\
MMLU-Pro & 58.6 $\pm$ 1.0 & 25.8 $\pm$ 2.3 & 16.3 $\pm$ 2.6 & 10 & 12.8 $\pm$ 6.1 & 10 / 0 \\
AIME & 28.5 $\pm$ 2.7 & 15.3 $\pm$ 2.6 & 9.5 $\pm$ 2.6 & 10 & 2.5 $\pm$ 2.4 & 3 / 0 \\
APPS & 39.5 $\pm$ 6.0 & 15.6 $\pm$ 4.9 & 3.9 $\pm$ 2.8 & 0 & $-$1.5 $\pm$ 2.4 & 0 / 0 \\
BigCodeBench & 11.0 $\pm$ 6.2 & 3.6 $\pm$ 2.3 & $-$1.8 $\pm$ 4.7 & 0 & 4.2 $\pm$ 2.7 & 1 / 0 \\
CodeContests & 34.8 $\pm$ 2.0 & 8.7 $\pm$ 4.5 & 2.1 $\pm$ 4.0 & 0 & $-$0.1 $\pm$ 3.3 & 0 / 0 \\
SuperGPQA & 59.0 $\pm$ 2.8 & 32.1 $\pm$ 5.8 & 5.8 $\pm$ 5.0 & 3 & 10.4 $\pm$ 7.6 & 5 / 0 \\
BIG-Bench Extra Hard & 37.6 $\pm$ 3.5 & 24.8 $\pm$ 5.7 & 1.0 $\pm$ 1.1 & 1 & 20.6 $\pm$ 5.3 & 10 / 0 \\
\bottomrule
\end{tabular}
\caption{\textbf{Ten random re-splits per dataset}, which, unlike the training sets, also reshuffle the test problems (original sizes; every probe and estimator refitted, configurations chosen on calibration). Mean $\pm$ s.d.\ over splits of cost saved (\%) relative to median pricing, and of our saving relative to the best external estimator in each split and to length from predicted success. ``Splits ahead'': splits in which ours is significantly ahead of every external estimator, and (last column) ahead of / behind length from predicted success. No external estimator is significantly cheaper than ours in any split. Values differ from \cref{tab:pools}; the conclusions do not.}
\label{tab:resplit}
\end{table}

\FloatBarrier
\section{Where the saving comes from}\label{app:subsets}
\begin{figure}[t]
\centering
\includegraphics[width=\linewidth]{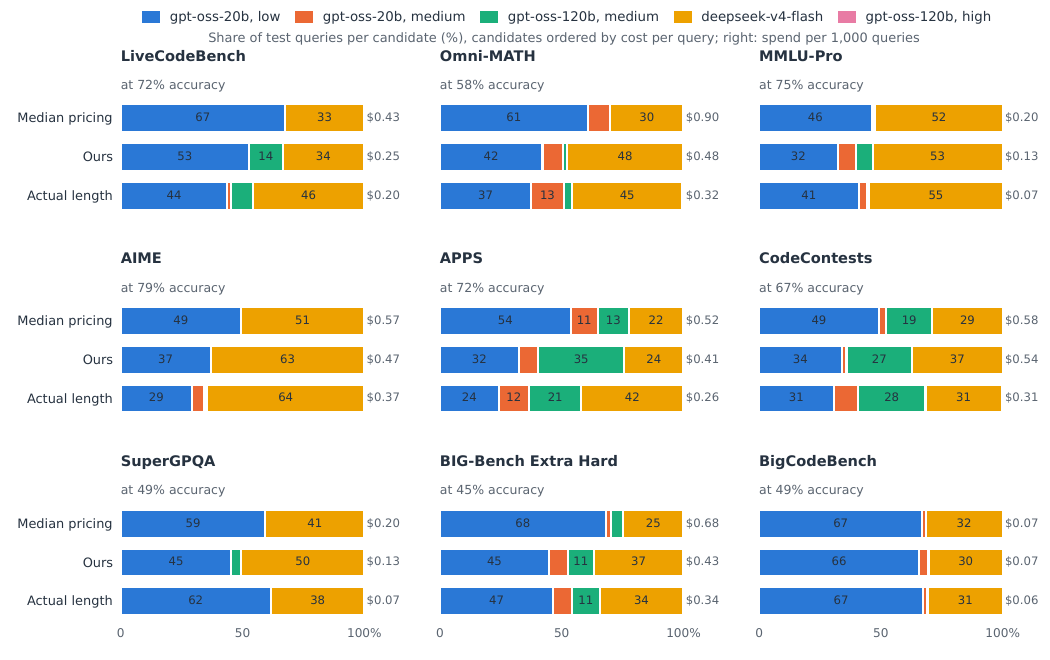}
\caption{\textbf{Candidate shares at matched accuracy} (one training set). Share of test queries (\%) sent to each candidate at the middle of the accuracy band that all three pricings reach; right: spend per 1,000 queries. Candidates are ordered by mean cost per query; shares below 10\% are unlabelled.}
\label{fig:shares}
\end{figure}

\paragraph{Where the queries go.}
On one training set, at the middle of each dataset's shared accuracy band, our router sends fewer queries than median pricing to gpt-oss-20b at low effort, the candidate with the lowest typical cost, and more to deepseek-v4-flash or gpt-oss-120b at medium effort (\cref{fig:shares}); on every dataset but MMLU-Pro, SuperGPQA and BigCodeBench, pricing at actual length shifts traffic the same way. The mix is not the whole story: on MMLU-Pro and SuperGPQA, pricing at actual length sends about the same shares as median pricing at a third of its spend, so its saving lies in which queries go to each candidate rather than how many. gpt-oss-120b at high effort receives at most 1\% of queries under any of the three. Per problem, 31--66\% change candidate relative to median pricing, mostly between gpt-oss-20b at low effort and deepseek-v4-flash: problems predicted to be short move to the more accurate deepseek-v4-flash (19--38\% of problems), and problems predicted to be long move to the cheap candidate (9--18\%).

\paragraph{Candidate subsets.}
We refit the router on every subset of the five candidates that leaves out one candidate, and on every pair (\cref{tab:subsets}; ten training sets). Leaving out the cheapest candidate, gpt-oss-20b at low effort, removes most of the saving, and leaving out deepseek-v4-flash removes less; leaving out either gpt-oss-120b effort or gpt-oss-20b at medium effort changes the saving by at most 2.1 points on LCB, Omni-MATH, MMLU-Pro, AIME, SuperGPQA and BIG-Bench Extra Hard. Choosing effort within one model is worth little: with only gpt-oss-20b's two efforts ours saves 20.2\% (LCB), 16.8\% (Omni-MATH) and 9.0\% (MMLU-Pro) relative to median pricing, with only gpt-oss-120b's 0.2\%, 7.0\% and 6.0\%. Among pairs, the saving lives in the cheap candidate paired with a strong candidate whose length varies widely. deepseek-v4-flash has the widest length spread of any candidate on all nine datasets (s.d.\ of log length 1.72 on LCB and Omni-MATH, against at most 1.40 and 1.47 for the other candidates), and on seven datasets its spread of cost per call exceeds the spread of prices between candidates.

\begin{table}[h]
\centering
\footnotesize
\setlength{\tabcolsep}{3pt}
\resizebox{\linewidth}{!}{%
\begin{tabular}{lrrrrrrrr}
\toprule
 & & \multicolumn{5}{c}{Without} & \multicolumn{2}{c}{Pair: gpt-oss-20b low and} \\
\cmidrule(lr){3-7}\cmidrule(lr){8-9}
Dataset & All five & 20b low & 20b med. & deepseek & 120b med. & 120b high & deepseek & 120b high \\
\midrule
LiveCodeBench & \textbf{27.8} & 6.5 & \textbf{28.4} & \textbf{22.4} & \textbf{27.8} & \textbf{28.3} & \textbf{29.0} & \textbf{32.1} \\
Omni-MATH & \textbf{31.5} & \textbf{16.2} & \textbf{32.6} & \textbf{15.2} & \textbf{31.8} & \textbf{32.0} & \textbf{33.8} & \textbf{22.4} \\
MMLU-Pro & \textbf{23.4} & -- & \textbf{25.5} & \textbf{17.0} & \textbf{24.0} & \textbf{23.2} & \textbf{27.0} & \textbf{25.9} \\
AIME & \textbf{14.9} & $-$0.2 & \textbf{14.9} & \textbf{7.0} & \textbf{15.0} & \textbf{14.9} & \textbf{15.0} & \textbf{10.7} \\
APPS & \textbf{16.8} & 1.3 & \textbf{15.0} & \textbf{11.5} & 11.6 & \textbf{19.3} & 16.3 & 9.1 \\
BigCodeBench & 0.4 & $-$45.0 & 1.8 & 3.8 & 0.8 & 1.8 & 2.9 & 6.2 \\
CodeContests & \textbf{9.4} & 4.3 & \textbf{10.1} & \textbf{7.1} & 9.0 & \textbf{10.2} & \textbf{10.7} & 11.2 \\
SuperGPQA & \textbf{33.0} & 0.7 & \textbf{33.4} & 15.4 & \textbf{34.0} & \textbf{32.8} & \textbf{34.6} & 24.1 \\
BIG-Bench Extra Hard & \textbf{23.5} & 3.8 & \textbf{21.9} & \textbf{26.2} & \textbf{22.9} & \textbf{24.3} & \textbf{20.9} & \textbf{33.5} \\
\bottomrule
\end{tabular}}
\caption{\textbf{Which candidates the saving needs.} Cost saved (\%) by our cost probe relative to median pricing, with the router refitted on a subset of the five candidates; bold: 95\% interval excludes zero. --: the frontiers do not overlap (MMLU-Pro without gpt-oss-20b at low effort, where deepseek-v4-flash has the lowest median cost per call).}
\label{tab:subsets}
\end{table}

\paragraph{Price gaps.} We move every candidate's per-token price toward or away from the candidates' geometric mean ($r_m^\alpha$, rescaled; $\alpha$ from 0, flat, to 5), keeping actual lengths fixed and recomputing costs from token counts (one training set; BigCodeBench is omitted, since its matched-accuracy band degenerates at some spreads). On LCB and Omni-MATH our saving grows with the price gap and saturates near the billed gap (3.1$\times$ between the most and least expensive candidate); on MMLU-Pro it is largest at flat prices. Beyond the billed gap the headroom shrinks: at a 29$\times$ spread it is 7--49\% across datasets, at 87$\times$ 2--42\%, and our saving falls with it (1--25\% at 87$\times$).

\FloatBarrier
\section{Three further model families}\label{app:families}
We added Qwen3-32B, GLM-4.7-Flash and Nemotron-3-Super-120B, all reasoning, on LCB, Omni-MATH and a 2,700-problem subset of MMLU-Pro (billed \$0.40--0.57 per million output tokens; Qwen3-32B's provider stops at 24,575 output tokens, and Nemotron has no valid answer for 27\% of Omni-MATH problems after retries), and fitted the same probes for them (test success AUC .72--.87, log-length $R^2$ .23--.81).

\paragraph{All eight candidates.} Averaged over ten training sets, with all eight candidates our router saves 28.8\%, 25.6\% and 23.0\% relative to median pricing, beats length from predicted success on MMLU-Pro (6.7 [0.3, 12.2]) and ties it on LCB and Omni-MATH, as with five candidates. In the full pool the new models are never needed (removing any one changes the saving by under 0.2 points on one training set): deepseek-v4-flash is cheaper than each of them at a higher accuracy.

\paragraph{Without deepseek-v4-flash and gpt-oss-120b.} We therefore also route in a pool of both gpt-oss-20b efforts and the three new models, without deepseek-v4-flash or gpt-oss-120b. There our router saves 29.5\% [17.3, 37.2], 12.4\% [$-$0.2, 21.7] and 22.7\% [9.9, 33.6] relative to median pricing on LCB, Omni-MATH and MMLU-Pro and ties length from predicted success; with all candidates except deepseek-v4-flash they save 25.3\%, 11.4\% and 22.4\%. Without deepseek-v4-flash no single candidate is cheap and accurate: the most accurate one, Nemotron-3-Super-120B (85.2\%, 75.2\% and 82.4\%), costs \$6.6, \$7.5 and \$2.1 per thousand calls against \$0.6, \$0.9 and \$0.2 for gpt-oss-20b at medium effort; our router's most accurate policy ends 1.6--3.2 points below it. With gpt-oss-120b in the pool, the router reaches Nemotron's accuracy on LCB at 64.8\% [42.3, 83.8] lower cost.

\FloatBarrier
\section{What drives length}\label{app:mechanism}
\Cref{tab:bracket} gives, per candidate, three test $R^2$ values of log mean output length: our probe's, which is the highest of every pre-generation estimator we evaluated (length from predicted success, ZeroRouter's pricing, prompt features, MixLLM-style and nearest-neighbour probes of the text encoder of \cref{app:baselines}, ridge, MixLLM-style and nearest-neighbour probes of Qwen3-Embedding-4B, Route-To-Reason, and on LCB the fine-tuned Qwen3-0.6B) on every candidate of every dataset; that of a linear map from the mean actual log length of the other four candidates on the same problem, which is not a deployable estimator but shows how much of a candidate's length is shared with other models; and the between-problem share, the share of the variance of a candidate's per-problem log length over test problems that is between problems rather than between samples (from the candidates' repeated samples; APPS has one sample per problem). Means over ten training sets.

\paragraph{What our probe predicts.}
On every candidate of every dataset our probe predicts length better than every other estimator (\cref{tab:bracket}) except one length per query, which, predicted from the same activations, matches or slightly exceeds it on some candidates (by at most .04) but saves less (\cref{tab:allest}), because it fixes the ratio of lengths between candidates for every query. Our $R^2$ comes closest to the between-problem share on LCB, Omni-MATH and BIG-Bench Extra Hard (e.g.\ LCB .71--.75 against .98--.99) and falls furthest below it on the knowledge datasets, for deepseek-v4-flash above all: on MMLU-Pro and SuperGPQA our probe explains .26 and .22 of its log length, other candidates' lengths explain .62 and .46, and the between-problem share is .92 and .96. Its length there depends on the question and is largely shared with other models, but our probe misses it. Predictors do worst on differences between candidates: across candidate pairs, our $R^2$ for the log length ratio is .06--.47 per dataset, against a between-problem share of .48--.85, and this $R^2$ of the length ratio is correlated with the share of a pair's headroom we capture (Spearman .49 over the 75 pairs with headroom above 5 points) better than per-candidate $R^2$ does (.18). On LCB (one training set), our probe predicts the shared length ($R^2$ .75) far better than the differences between candidates (.41). Still, pricing with the predicted shared length alone keeps only 68\% of the full probe's saving on LCB, 62\% on MMLU-Pro and 42--45\% on Omni-MATH and AIME (ten training sets): the differences it does predict are worth a third or more of the saving.

\begin{table}[h]
\centering
\footnotesize
\setlength{\tabcolsep}{3pt}
\resizebox{\linewidth}{!}{%
\begin{tabular}{lccccc}
\toprule
Dataset & 20b low & 20b med. & deepseek & 120b med. & 120b high \\
\midrule
LiveCodeBench & .74 / .85 / .99 & .71 / .96 / .99 & .72 / .92 / .99 & .73 / .95 / .99 & .75 / .94 / .98 \\
Omni-MATH & .68 / .60 / .84 & .66 / .86 / .89 & .71 / .79 / .95 & .72 / .87 / .94 & .73 / .88 / .95 \\
MMLU-Pro & .61 / .42 / .71 & .52 / .68 / .79 & .26 / .62 / .92 & .62 / .76 / .88 & .45 / .67 / .89 \\
AIME & .41 / .52 / .89 & .44 / .85 / .93 & .50 / .79 / .97 & .49 / .86 / .92 & .51 / .87 / .94 \\
APPS & .56 / .69 / -- & .54 / .80 / -- & .63 / .86 / -- & .63 / .88 / -- & .68 / .88 / -- \\
BigCodeBench & .69 / .66 / .97 & .61 / .86 / .96 & .47 / .71 / .95 & .71 / .85 / .97 & .45 / .67 / .88 \\
CodeContests & .46 / .67 / .92 & .35 / .83 / .93 & .52 / .80 / .96 & .42 / .88 / .94 & .43 / .86 / .95 \\
SuperGPQA & .75 / .50 / .94 & .47 / .64 / .88 & .22 / .46 / .96 & .63 / .77 / .92 & .34 / .54 / .90 \\
BIG-Bench Extra Hard & .56 / $-$.01 / .81 & .81 / .82 / .89 & .83 / .72 / .96 & .90 / .75 / .92 & .84 / .69 / .94 \\
\bottomrule
\end{tabular}}
\caption{\textbf{How much of each candidate's length we predict, and how much depends on the problem.} Test $R^2$ of log mean output length: our probe / a linear map from the other candidates' actual lengths on the same problem / the share of variance between problems rather than between samples. A dash means the share is not computable, because APPS has one sample per problem.}
\label{tab:bracket}
\end{table}
\paragraph{The success probes carry more than difficulty.}
Where length from predicted success ties the cost probe (LCB, Omni-MATH), it does not price by difficulty alone. Empirical difficulty, a problem's measured solve rate over the five candidates (with its square), is known only after running and grading all of them, and it is a weak price even in-domain (one training set). It saves 21.5\% on LCB, against 34.4\% for length from predicted success, because the logits carry length cues beyond the solve rate. Among LCB problems that every other candidate always solves, problems with a lower predicted success logit have longer outputs (Spearman $-$.65 to $-$.85).

\paragraph{Empirical difficulty and the cost probe.} Here empirical difficulty is a problem's solve rate across the other four candidates, so it never shares samples with the length it explains. Per candidate, we fit log length on the calibration problems from the cost probe's prediction, from empirical difficulty (with its square), or from both, and score the test problems (one training set). Empirical difficulty alone explains a median $R^2$ of .57 on AIME, .54 on LCB, .45 on CodeContests, .38 on APPS and .33 on Omni-MATH, and .17 or less on BIG-Bench Extra Hard, MMLU-Pro, BigCodeBench and SuperGPQA. Adding it to the cost probe's prediction raises $R^2$ by a median of .16 on AIME and .13 on CodeContests, and by .05 or less elsewhere.

\FloatBarrier
\section{Generalisation}\label{app:transfer}
\subsection{Settings}\label{app:settings}
\paragraph{Few labels.} Each cost estimator is refitted on $n$ random training problems, two subsets in each of the ten training sets, with the same success probes for every estimator; savings are relative to median pricing fitted on the same $n$ problems. ZeroRouter's pricing (its item-response latent, the map to it from the scout's activations, and its bin lookup) is fitted on the $n$ problems; the $k$NN neighbour count and ZeroRouter's dimension and bin count are chosen on calibration.
\paragraph{Onboarding.} The held-out model is gpt-oss-20b (both efforts), gpt-oss-120b (both efforts) or deepseek-v4-flash, so that no sibling effort remains in the pool, with four random choices of the $k$ labelled problems in each training set, the same for every method. ZeroRouter's onboarding fits the new model's ability on the $k$ labelled problems in an item-response latent fitted on the remaining candidates and fills its per-bin length table from them~\citep{zerorouter}; the CARROT-style method averages those problems' outcomes and lengths over nearest neighbours. ZeroRouter's dimension and bin count and the $k$NN neighbour count are chosen on calibration.
\paragraph{Transfer.} The domains are coding (LCB, APPS, BigCodeBench, CodeContests) and math and knowledge (Omni-MATH, MMLU-Pro, AIME, SuperGPQA, BIG-Bench Extra Hard). Every choice (regularization, Platt scaling, neighbour counts, latent dimension, bin count) is made on the source datasets' calibration splits, and each training set reshuffles every dataset's splits with the same seed. External estimators are transferred the same way, each with its own success model (MixLLM-style: a logistic probe on the text embedding; $k$NN; ZeroRouter: its latent and bins). Omni-MATH enters as its 500-problem set (test 150), whose scout, the Instruct model, matches the other datasets'. BigCodeBench and CodeContests, where no in-domain estimator saves significantly, are not used as targets: there the transferred success probes are badly miscalibrated (BigCodeBench log loss 1.17 against .67 in-domain) and the apparent gains are not robust. Transferred success predictions are weaker than in-domain ones (LCB AUC .67 against .83), so the transferred router's saving comes from its cost probe.
\paragraph{Mixed traffic.} Training and calibration problems are the union of each dataset's splits; the test set has 150 problems from each dataset (1,350); Omni-MATH enters as its 500-problem set. Both probes are refitted on the mixture, and bootstrap resamples are drawn within each benchmark.

\subsection{Results}
\begin{figure}[h]
\centering
\includegraphics[width=\linewidth]{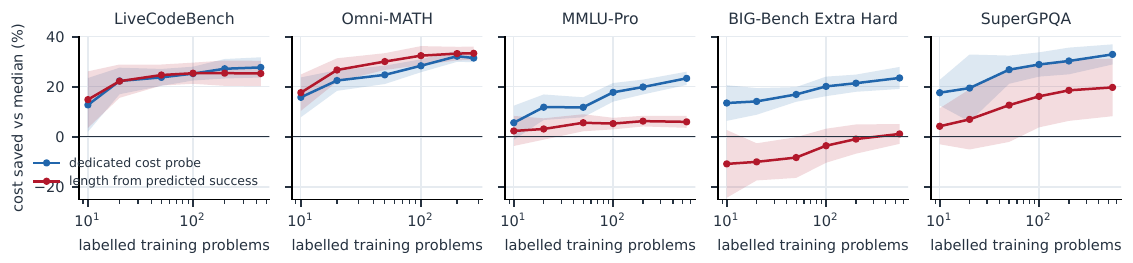}
\caption{\textbf{Length labels needed.} Cost saved relative to median pricing fitted on the same problems, when cost is fitted on $n$ labelled training problems: our cost probe (blue) versus length from predicted success (red), which needs length labels only to map success logits to length. Two random subsets per training set; bands: $\pm$1 s.d.\ over training sets.}
\label{fig:labels}
\end{figure}
\paragraph{Few labels.} Ours exceeds every external estimator with paired intervals excluding zero on LCB and Omni-MATH from $n$=20 (except Omni-MATH at $n$=50), on MMLU-Pro from $n$=100 and on SuperGPQA from $n$=200. On BIG-Bench Extra Hard the text estimators stay within noise of ours at every $n$; below 100 problems ours leads on average (by 16.3 points over MixLLM-style embeddings at 50), with intervals that include zero. Labelling one sampled output per training problem instead of all of them gives up 1--3 points (one training set).

\begin{table}[h]
\centering
\footnotesize
\setlength{\tabcolsep}{4pt}
\begin{tabular}{lrrrrr}
\toprule
 & LCB & Omni-MATH & MMLU-Pro & SuperGPQA & BBEH \\
\midrule
Ours, $k$=10 & 19.5 & 22.7 & 19.2 & 26.7 & 9.7 \\
ZeroRouter onboarding, $k$=10 & 7.3 & 17.4 & 9.8 & 13.7 & 3.4 \\
CARROT-style $k$NN, $k$=10 & 5.2 & 11.4 & 9.5 & 13.8 & 8.3 \\
Median length + base rate, $k$=10 & $-$13.6 & 7.6 & 2.1 & 9.0 & 1.8 \\
\midrule
Ours $-$ ZeroRouter, $k$=10 & 12.2 [$-$5.9, 26.7] & 5.2 [$-$7.9, 18.2] & 9.4 [$-$0.9, 19.3] & 13.0 [$-$1.8, 25.9] & 6.3 [$-$11.5, 29.1] \\
Ours $-$ ZeroRouter, $k$=50 & 9.6 [3.0, 18.7] & 5.1 [$-$5.4, 17.7] & 8.1 [$-$0.5, 16.5] & 12.8 [4.8, 20.8] & 4.8 [$-$1.5, 10.5] \\
\midrule
\emph{All candidates on full probes} & 27.8 & 31.5 & 23.4 & 33.0 & 23.5 \\
\bottomrule
\end{tabular}
\caption{\textbf{Onboarding a new model from $k$ examples.} Cost saved (\%) relative to the full pool at median pricing, mean over three held-out models (gpt-oss-20b with both efforts, gpt-oss-120b with both efforts, deepseek-v4-flash) and four random choices of the $k$ labelled problems per training set (the same problems for every method). Differences in points, with 95\% intervals.}
\label{tab:onboard}
\end{table}
\paragraph{Combining source and target labels.} The remaining results in this appendix are on one training set. Combining both sources of labels is better than either: with ten labelled target problems added to the source data, the cost probe saves 27.9\% on MMLU-Pro and 13.1\% on AIME (with the target's success probes), against 8.1\% and 4.4\% from the ten problems alone, and 24.1\% against 14.7\% on BIG-Bench Extra Hard with twenty. 
\paragraph{Transferring length from predicted success.} With the target's own success probes and a map trained on the other datasets, length from predicted success keeps its saving where the target's kind of difficulty resembles its sources' (APPS 17.1\%, AIME 12.3\%, CodeContests 9.3\%) and collapses elsewhere: from 34.4\% to $-$6.2\% on LCB, 19.6\% to $-$21.9\% on SuperGPQA, 18.0\% to 6.5\% on Omni-MATH and 5.5\% to $-$20.5\% on BIG-Bench Extra Hard, against 30.2\%, 17.7\%, 15.7\% and 16.1\% for the cost probe. Bootstrapping the drop from in-domain to transferred pricing with the target's success probes fixed, ours loses significantly less than length from predicted success on LCB, SuperGPQA and BIG-Bench Extra Hard, and on no dataset significantly more.
\paragraph{Why the cost probe transfers.}
Part of what transfers is shared structure: 24--40\% of each candidate's variance in log length lies between benchmarks, and the same benchmarks are long for every candidate (mean correlation of benchmark means across candidates .84). What does not transfer is the map from difficulty to length. Matching the problems of two benchmarks by empirical difficulty (measured solve rate per candidate, five bins), their lengths still differ by 1.14 log units on average ($\times$3.1 over 470 benchmark-pair and difficulty-bin groups). Our cost probe, trained on neither benchmark, predicts these gaps (correlation .95, 90\% of the squared gap), and length from predicted success 30\%. Routing also needs less than calibrated lengths: on MMLU-Pro length $R^2$ falls sharply under transfer (.02--.46 against .29--.63) while routing transfers fully.

\FloatBarrier
\section{Representation, fine-tuning and scout size}\label{app:size}
\begin{table}[h]
\centering
\footnotesize
\setlength{\tabcolsep}{4pt}
\begin{tabular}{lrrrrrr}
\toprule
 & \multicolumn{3}{c}{Ours vs.\ embedding cost probe} & Ours vs.\ & \multicolumn{2}{c}{Success AUC} \\
\cmidrule(lr){2-4}\cmidrule(lr){6-7}
Dataset & Ridge & MixLLM & $k$NN & text router & text & ours \\
\midrule
LiveCodeBench & \textbf{16.9} & \textbf{19.5} & \textbf{23.1} & \textbf{12.0} & .75 & .83 \\
Omni-MATH & \textbf{15.8} & \textbf{18.8} & \textbf{19.6} & \textbf{18.5} & .73 & .83 \\
MMLU-Pro & \textbf{15.8} & \textbf{13.3} & \textbf{12.0} & \textbf{12.6} & .64 & .72 \\
AIME & \textbf{9.7} & \textbf{10.1} & \textbf{12.2} & \textbf{6.8} & .65 & .78 \\
APPS & 6.5 & 11.7 & 9.0 & 5.0 & .75 & .84 \\
BigCodeBench & 1.2 & 2.0 & 0.8 & 4.3 & .61 & .64 \\
CodeContests & 1.3 & 3.0 & 5.7 & $-$4.7 & .65 & .74 \\
SuperGPQA & 7.0 & 10.3 & \textbf{13.3} & 7.0 & .54 & .57 \\
BIG-Bench Extra Hard & 5.4 & 4.1 & 4.6 & 8.4 & .76 & .79 \\
\bottomrule
\end{tabular}
\caption{\textbf{A text-embedding model of the scout's size.} Cost saved (\%) by our router relative to routers that predict cost from Qwen3-Embedding-4B (our ridge recipe, a MixLLM-style ensemble, CARROT-style $k$NN; all with our success probes) and to a fully text-based router that also predicts success from the embedding. Bold: 95\% interval excludes zero. Success AUC: mean over candidates on the test set.}
\label{tab:textemb}
\end{table}
\paragraph{Representation, probe and scout size.}
\Cref{tab:textemb} predicts cost from Qwen3-Embedding-4B~\citep{qwen3emb} with our ridge recipe, a MixLLM-style ensemble and CARROT-style $k$-nearest neighbours, each with our success predictions, and builds a fully text-based router that also predicts success from the embedding with our logistic recipe. We do not charge the embedding's own encoder pass, which would only favour our cost probe. On PCA-reduced features of our scout the three probes are within about 7 points of one another (one training set), so the gain comes from the representation rather than the regressor. \Cref{tab:size} varies the scout, including three other families (Phi-4-mini~\citep{phi4mini}, Granite-3.3-2B, SmolLM2-1.7B~\citep{smollm2}), while keeping prompts, layers and both probes fixed (ten training sets). Any small scout works: size and release change the saving by a few points, and none beats ours on average.

\paragraph{Fine-tuning.}
We fully fine-tuned DistilBERT 66M, jina-embeddings 137M and Qwen3-0.6B on the same training problems and length labels (learning rate and epoch chosen on calibration error) and used each as a cost estimator with our success predictions. On one training set ours saves 11.9--25.6\% relative to each on Omni-MATH and MMLU-Pro and 11.7--14.9\% relative to the two smaller encoders on LCB; averaged over ten training sets it saves 6.5\% [$-$1.2, 12.2] relative to the fine-tuned Qwen3-0.6B on LCB. The same Qwen3-0.6B, frozen, with a linear probe on its activations, saves 31.9\% and 22.0\% relative to median pricing on Omni-MATH and MMLU-Pro (one training set), more than its fine-tuned copy (20.0\% and 19.3\%), despite pairing with its own weaker success probes.

\paragraph{The prefill router's success pipeline.}
The prefill router~\citep{prefill} selects a layer, a pooling and a PCA dimension per target by cross-validation over the encoder's upper layers, concatenates the selected features and trains an ensemble of shared-trunk MLPs. We reimplemented this pipeline on the same encoder, splits and cost probe (one training set). The two tie on LCB, MMLU-Pro and Omni-MATH: ours saves $-$0.1\% [$-$7.7, 4.0], 1.1\% [$-$1.0, 3.0] and $-$2.2\% [$-$5.7, 1.9] relative to it. Ours saves 8.4\% [$-$2.5, 18.4] relative to it on BIG-Bench Extra Hard, and it is ahead by 6.0\% [0.4, 11.2] on SuperGPQA, where success is barely predictable for either (AUC .58 and .57).

\begin{table}[t]
\centering
\footnotesize
\setlength{\tabcolsep}{3pt}
\resizebox{\linewidth}{!}{%
\begin{tabular}{lrrrrrr}
\toprule
 & \multicolumn{3}{c}{Omni-MATH} & \multicolumn{3}{c}{MMLU-Pro} \\
\cmidrule(lr){2-4}\cmidrule(lr){5-7}
Scout & $R^2$ & vs.\ median & Ours vs.\ it & $R^2$ & vs.\ median & Ours vs.\ it \\
\midrule
Qwen3-0.6B & .61 & 27.9 & 7.3 [$-$0.6, 15.7] (5) & .33 & 17.0 & \textbf{8.2 [0.5, 14.0]} (8) \\
Qwen3-1.7B & .65 & 29.0 & 4.1 [$-$5.5, 13.4] (4) & .40 & 19.4 & 5.2 [$-$1.5, 12.1] (7) \\
Qwen3-4B (original) & .66 & 28.1 & 3.6 [$-$6.5, 12.3] (4) & .44 & 20.6 & 3.4 [$-$2.5, 7.4] (8) \\
\midrule
Phi-4-mini (3.8B) & .64 & 28.2 & 7.6 [$-$2.4, 16.5] (7) & .43 & 21.4 & 2.0 [$-$6.2, 6.8] (5) \\
Granite-3.3 (2.5B) & .62 & 25.1 & 5.6 [$-$3.3, 14.6] (3) & .38 & 18.2 & \textbf{7.1 [2.1, 12.8]} (10) \\
SmolLM2 (1.7B) & .58 & 25.2 & 5.9 [$-$3.7, 14.0] (6) & .33 & 16.9 & \textbf{7.5 [2.1, 11.5]} (9) \\
\midrule
Qwen3-4B-2507 (ours) & .70 & 31.5 & -- & .49 & 23.4 & -- \\
\bottomrule
\end{tabular}}
\caption{\textbf{Other scouts.} Each scout's activations feed both probes, refitted with the paper's recipes (Omni-MATH and MMLU-Pro, billed costs). $R^2$: test log-length $R^2$, mean over candidates. vs.\ median: cost saved (\%) relative to median pricing with the row's own success predictions. Ours vs.\ it: cost saved by our router relative to the row's, with 95\% intervals (bold: excludes zero) and the number of training sets in which ours is significantly ahead; ours is significantly behind in one training set only (Phi-4-mini, MMLU-Pro).}
\label{tab:size}
\end{table}

\FloatBarrier
\section{Deployment}\label{app:deploy}
\paragraph{Routing near the top.}
The router's most accurate policy ties the most accurate candidate on AIME, is 2.0 points above it on BIG-Bench Extra Hard and below it elsewhere, and where it reaches deepseek-v4-flash's accuracy it spends as much as calling it alone (AIME, SuperGPQA, BIG-Bench Extra Hard) or more (the coding datasets). Giving up one point of accuracy saves 5.1\% on AIME and 9.4\% on SuperGPQA (intervals excluding zero) and nothing significant elsewhere, while giving up five points saves 14.9--40.3\% on the six datasets other than APPS, BigCodeBench and CodeContests. On the same six, the router is 45--83\% cheaper than gpt-oss-20b at medium effort at its accuracy and, where it reaches its accuracy, 50--92\% cheaper than gpt-oss-120b at high effort.

\paragraph{Deployable policies.}
Our main metric compares cost--accuracy frontiers on the test set; a deployed router must fix its trade-off $V$ in advance. For each estimator and each of ten accuracy targets spanning the band every estimator reaches on the calibration split, we choose the cheapest $V$ that reaches the target on calibration and apply it once to the test set. We compare its test cost with median pricing's deployed policies, interpolated at the accuracy it actually reaches (\cref{tab:deploy}; ten training sets). Our deployed policies save 19.6--37.9\% on LCB, Omni-MATH, MMLU-Pro, AIME, SuperGPQA and BIG-Bench Extra Hard; they cost 1.0--3.3\% more than our own test frontier at the accuracy they reach and miss their target by 2.2--5.3 points. They lead the best deployed external estimator on LCB, Omni-MATH and MMLU-Pro with intervals excluding zero in every training set, on AIME in 9 and SuperGPQA in 7 of 10, and tie it on BIG-Bench Extra Hard; they lead length from predicted success on MMLU-Pro and BIG-Bench Extra Hard in every training set and on SuperGPQA in 7. Deployed savings exceed the frontier savings mainly because median pricing deploys less well: its calibration-chosen policies fall 8.6--16.5\% short of its own test frontier on MMLU-Pro, AIME, APPS, BigCodeBench, CodeContests and SuperGPQA, against at most 9.4\% for ours. On APPS, BigCodeBench and CodeContests our lead is within noise.

\paragraph{The router's own cost.}
The features come from one forward pass of Qwen3-4B over the prompt: 127--1,216 tokens on average per dataset, 19--28\,ms median latency on one H100 at batch size one. A router that predicts success from the prefill already incurs the cost of this pass, so the cost probe adds nothing to it. Against no routing at all, the pass is not free when the routed models are as cheap as ours (well under a dollar per thousand queries). Self-hosted at an assumed \$2 per H100-hour with unoptimised batched inference, it costs 2--10\% of our deployed saving on Omni-MATH, AIME, MMLU-Pro and SuperGPQA and about 30\% on BIG-Bench Extra Hard and APPS, whose prompts are long; at the input price of the smallest Qwen3 model listed on OpenRouter (14B, \$0.12 per million tokens), it costs 6--51\% of the saving, and more than the saving on APPS. Its share falls as the routed models get more expensive.

\paragraph{Per-query budgets.}
We cap the cost of every call at $X$ (a call that would exceed it is cut off and fails), with $X$ at the 50th, 70th, 80th, 90th and 95th percentile of deepseek-v4-flash's cost per call on the training and calibration problems, and send each query to the candidate with the highest predicted probability of succeeding within the cap: the success probe times the chance that the candidate's length fits, from the predicted length and the training residuals (ten training sets, full test sets). At the tightest cap the router beats deepseek-v4-flash alone on LCB (8.2 points of accuracy [5.8, 10.8]) and MMLU-Pro (7.0 [5.7, 8.3]), and at looser caps it ties it or falls 1--3 points behind. Knowing each problem's actual length would beat it by 9.3--15.5 points at the tightest cap, so there is room under budgets, but a probe trained directly on whether a call fits predicts it no better (within 2 points of ours everywhere). Against median pricing under the same caps our probe is ahead by up to 12.6 points (LCB at the 90th percentile) and behind significantly at no cap.

\begin{table}[t]
\centering
\footnotesize
\setlength{\tabcolsep}{3.5pt}
\begin{tabular}{lrlrrr}
\toprule
 & Ours vs median, & Best external & Ours $-$ best & Ours $-$ length & Ours: frontier loss \\
Dataset & deployed & (deployed) & external & from pred.\ success & / target miss \\
\midrule
LiveCodeBench & 31.3 (7.0) & MixLLM 14.9 & 16.3 (10) & 2.7 (2) & 2.4 / 2.6 \\
Omni-MATH & 35.0 (2.2) & ZeroRouter 13.9 & 21.5 (10) & $-$1.6 (0) & 2.0 / 5.3 \\
MMLU-Pro & 31.1 (4.8) & $k$NN 18.1 & 12.7 (10) & 17.3 (10) & 1.0 / 2.3 \\
AIME & 19.6 (2.1) & ZeroRouter 10.3 & 9.4 (9) & 2.2 (2) & 1.3 / 4.6 \\
APPS & 23.3 (5.1) & ZeroRouter 15.8 & 5.2 (1) & $-$2.4 (0) & 9.4 / 3.6 \\
BigCodeBench & 7.0 (5.2) & $k$NN 5.7 & 2.3 (0) & 3.8 (1) & 9.2 / 2.2 \\
CodeContests & 13.5 (4.4) & $k$NN 8.4 & 5.7 (2) & $-$0.3 (0) & 6.3 / 5.6 \\
SuperGPQA & 37.9 (7.1) & MixLLM 24.6 & 14.1 (7) & 15.4 (7) & 3.3 / 2.5 \\
BIG-Bench Extra Hard & 24.0 (4.5) & MixLLM 24.5 & $-$1.0 (0) & 28.7 (10) & 2.2 / 2.2 \\
\bottomrule
\end{tabular}
\caption{\textbf{Deployable policies.} Each estimator picks on the calibration split the cheapest $V$ reaching each of ten accuracy targets, applied once to the test set. Cost saved (\%) relative to median pricing's deployed policies at the accuracy reached, mean over targets (in parentheses: s.d.\ over training sets for our saving; for differences, the number of training sets in which ours is significantly ahead; ours is significantly behind only in one APPS and one Omni-MATH set, against length from predicted success). Last column: extra cost over our own test frontier at the accuracy reached (\%) / mean $|$test accuracy $-$ target$|$ (points). Billed prices.}
\label{tab:deploy}
\end{table}

\FloatBarrier
\section{Software engineering}\label{app:agents}
On 1,432 SWE-Smith instances~\citep{swesmith}, each of the five candidates writes one patch (deepseek-v4-flash pinned as elsewhere), labelled by executing the repository's tests in a sandbox. Pass rates are low and close together (25--37\%), but which instances are solvable is predictable from the issue (success AUC .80--.83 per candidate on the 389 test instances where every candidate returned a patch). Output length varies 5--28$\times$ between a candidate's 10th- and 90th-percentile instances and is only modestly predictable (log-length $R^2$ .09--.37). Over ten training sets, per-query cost knowledge is worth 25.4\% [9.8, 35.5] over median pricing, and our cost probe saves 7.3\% [$-$4.7, 17.8], a capture of .29. No estimator differs significantly from ours: we save 5.7\% [$-$1.6, 11.9] relative to mean pricing, 3.6\% [$-$7.0, 13.5] relative to the prompt-feature GBM, 5.5\% [$-$3.8, 14.5] relative to MixLLM-style embeddings and 2.2\% [$-$4.8, 14.1] relative to length from predicted success, and ZeroRouter's whole router is within noise of ours ($-$5.9\% [$-$22.5, 11.6] with the 4B scout). One-shot patch generation thus behaves like the coding datasets where routing pays least: the headroom is real, but little of it is captured.

\FloatBarrier
\end{document}